%% file: manifold_GPLVM.tex
\documentclass{article}
\input{header}

\begin{document}

\maketitle

\input{abstract}

\input{intro}

\input{background}

\input{results}

\input{discussion}

\input{acknowledgements}

\normalsize
\bibliographystyle{apalike}
\bibliography{references}

\clearpage

\begin{appendices}
\crefalias{section}{appendix}
\input{appendix}
\end{appendices}

\bibliographystyleAPP{apalike}
\bibliographyAPP{appendix}

\end{document}

%% file: header.tex
\usepackage[compress]{natbib}
\usepackage[resetlabels]{multibib}
\newcites{APP}{References}

\usepackage[utf8]{inputenc} 
\usepackage[T1]{fontenc}    

\usepackage[table]{xcolor}
\definecolor{dred}{rgb}{0.6,0,0}
\definecolor{dpurple}{HTML}{A020F0}
\definecolor{dblue}{rgb}{0,0,0.6}
\definecolor{hlcolor}{rgb}{1,1,0.8}
 
\usepackage[colorlinks=true, urlcolor = dblue, linkcolor=dblue, citecolor=dred]{hyperref}       

\usepackage{graphicx}
\usepackage{caption}
\usepackage{subcaption}
\usepackage{amsmath,amssymb,bm}

\usepackage[nameinlink]{cleveref}
\usepackage{url}            
\usepackage{booktabs}       
\usepackage{amsfonts}       
\usepackage{nicefrac}       
\usepackage{microtype}      
\usepackage{soul}
\usepackage{authblk}
\usepackage{lmodern}
\usepackage[margin=1in]{geometry}

\usepackage[page]{appendix}

\sethlcolor{hlcolor}
\Crefname{equation}{Equation}{Equations}
\Crefname{figure}{Figure}{Figures}
\creflabelformat{equation}{#2#1#3}
\crefrangelabelformat{equation}{#3#1#4-#5#2#6}
 
\renewcommand\b\bm
\newcommand\Exp{\text{Exp}}
\title{Manifold GPLVMs for discovering\\ non-Euclidean latent structure in neural data \vspace{0.5em}}

\author[1]{\normalsize \bfseries Kristopher T.\ Jensen${}^{\rm @}$}
\author[1]{\bfseries Ta-Chu Kao}
\author[2]{\bfseries Marco Tripodi}
\author[1]{\bfseries Guillaume Hennequin}
\affil[1]{ \small Computational and Biological Learning Lab, Department of Engineering, University of Cambridge, Cambridge, UK}
\affil[2]{ \small MRC Laboratory of Molecular Biology, Neurobiology Division, Cambridge, UK}
\date{\vspace*{-1em}
\normalsize ${}^{\rm @}$ Corresponding author (ktj21@cam.ac.uk)}

%% file: abstract.tex
\begin{abstract}\noindent
A common problem in neuroscience is to elucidate the collective neural representations of behaviorally important variables such as head direction, spatial location, upcoming movements, or mental spatial transformations.
Often, these latent variables are internal constructs not directly accessible to the experimenter.
Here, we propose a new probabilistic latent variable model to simultaneously identify the latent state and the way each neuron contributes to its representation in an unsupervised way.
In contrast to previous models which assume Euclidean latent spaces, we embrace the fact that latent states often belong to symmetric manifolds such as spheres, tori, or rotation groups of various dimensions.
We therefore propose the manifold Gaussian process latent variable model (mGPLVM), where neural responses arise from (i) a shared latent variable living on a specific manifold, and (ii) a set of non-parametric tuning curves determining how each neuron contributes to the representation.
Cross-validated comparisons of models with different topologies can be used to distinguish between candidate manifolds, and variational inference enables quantification of uncertainty.
We demonstrate the validity of the approach on several synthetic datasets, as well as on calcium recordings from the ellipsoid body of \textit{Drosophila melanogaster} and extracellular recordings from the mouse anterodorsal thalamic nucleus.
These circuits are both known to encode head direction, and mGPLVM correctly recovers the ring topology expected from neural populations representing a single angular variable.
\end{abstract}

%% file: intro.tex
\section{Introduction}

The brain uses large neural populations to represent low-dimensional quantities of behavioural relevance such as location in physical or mental spaces, orientation of the body, or motor plans.
It is therefore common to project neural data into smaller latent spaces as a first step towards linking neural activity to behaviour~\citep{Cunningham2014}.
This can be done using a variety of linear methods such as PCA or factor analysis~\citep{cunningham2015linear}, or non-linear dimensionality reduction techniques such as tSNE~\citep{maaten2008visualizing}.
Many of these methods are explicitly probabilistic, with notable examples including GPFA~\citep{Yu2009} and LFADS~\citep{Pandarinath2018}.
However, all these models project data into Euclidean latent spaces, thus failing to capture the inherent non-Euclidean nature of variables such as head direction or rotational motor plans~\citep{Seelig2015, Chaudhuri2019, Finkelstein2015,Wilson2018}.

\input{_concept.tex}

Most models in neuroscience justifiably assume that neurons are smoothly tuned~\citep{Stringer2019}.
As an example, a population of neurons representing an angular variable $\theta$ would respond similarly to some $\theta$ and to $\theta+\epsilon$ (for small $\epsilon$). 
While it is straigthforward to model such smoothness by introducing smooth priors for response functions defined over $\mathbb{R}$, the activity of neurons modelled this way would exhibit a spurious discontinuity as the latent angle changes from $2\pi$ to $0+\epsilon$.
We see that appropriately modelling smooth neuronal representations requires keeping the latent variables of interest on their natural manifold (here, the circle), instead of an ad-hoc Euclidean space.
While periodic kernels have commonly been used to address such problems in GP regression \citep{MacKay1998}, topological structure has not been incorporated into GP-based latent variable models due to the difficulty of doing inference in such spaces.

Here, we build on recent advances in non-Euclidean variational inference \citep{Falorsi2019} to develop the manifold Gaussian process latent variable model (mGPLVM), an extension of the GPLVM framework \citep{Lawrence2005,Titsias2010a,Wu2017,Wu2018} to non-Euclidean latent spaces including tori, spheres and $SO(3)$ (\Cref{fig:concept}).
mGPLVM jointly learns the fluctuations of an underlying latent variable $g$ \emph{and} a probabilistic ``tuning curve'' $p(f_i | g)$ for each neuron $i$.
The model therefore provides a fully unsupervised way of querying how the brain represents its surroundings and a readout of the relevant latent quantities.
Importantly, the probabilistic nature of the model enables principled model selection between candidate manifolds. 
We provide a framework for scalable inference and validate the model on both synthetic and experimental datasets.

%% file: _concept.tex
\begin{figure}[!t]
    \centering
    \includegraphics[width = 0.9\textwidth, trim={0 0 0 0}, clip=true]{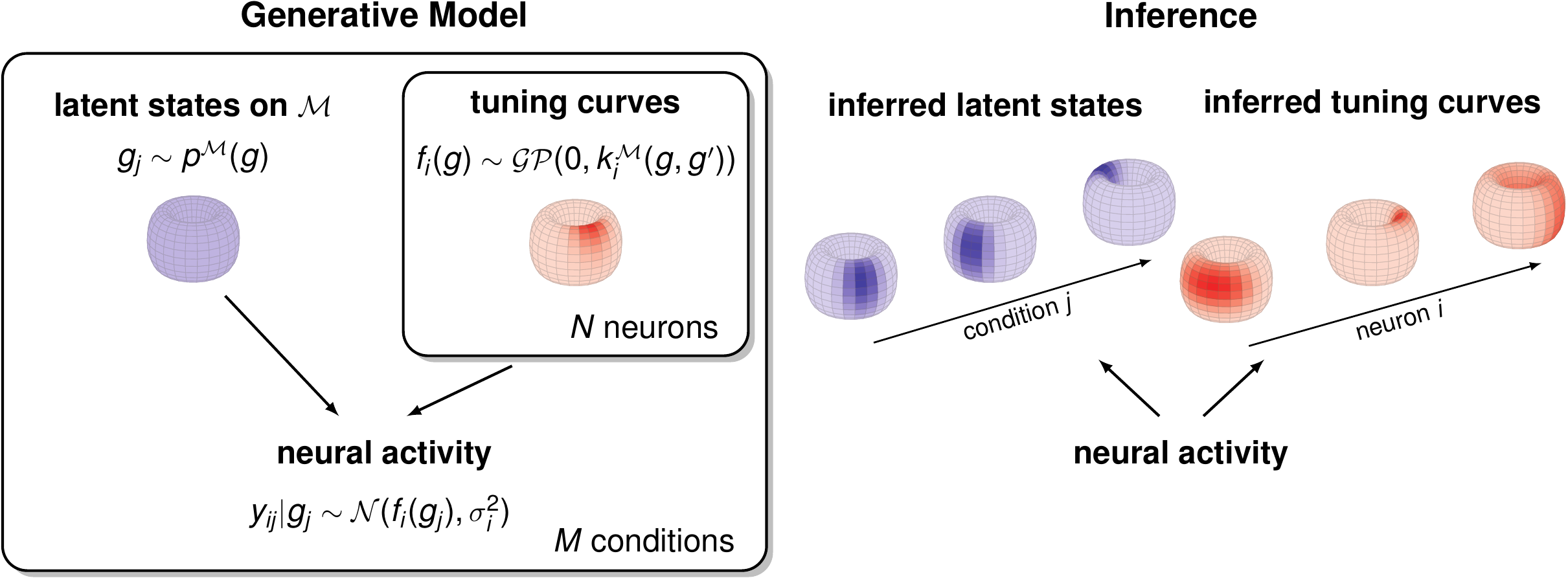}
    \caption{\label{fig:concept}
        {\bfseries Schematic illustration of the manifold Gaussian process latent variable model (\mbox{mGPLVM}).}
        In the generative model (left), neural activity arises from (i) $M$ latent states $\{g_j\}$ on a manifold $\mathcal{M}$, each corresponding to a different condition $j$ (e.g.\ time or stimulus), and (ii) the tuning curves of $N$ neurons, modelled as Gaussian processes and sharing the same latent states $\{g_j \}$ as inputs.
        Using variational inference, mGPLVM jointly infers the global latent states and the tuning curve of each neuron on the manifold (right).}
    \vspace*{-1em}
\end{figure}

%% file: background.tex
\section{Manifold Gaussian process latent variable model}

The main contribution of this paper is mGPLVM, a Gaussian process latent variable model \citep{Titsias2010a,Wu2018} defined for non-Euclidean latent spaces.
We first present the generative model (\Cref{subsec:generative}), then explain how we perform approximate inference using reparameterizations on Lie groups (\citealp{Falorsi2019}; \Cref{subsec:inference}).
Lie groups include Euclidean vector spaces $\mathbb{R}^n$ as well as other manifolds of interests to neuroscience such as tori $T^n$ \citep{Chaudhuri2019, rubin2019revealing} and the special orthogonal group $SO(3)$ (\citealp{Wilson2018, Finkelstein2015}; extensions to non-Lie groups are discussed in \Cref{subsec:spheres}).
We then provide specific forms for variational densities and kernels on tori, spheres, and $SO(3)$ (\Cref{subsec:tori_and_so3}).
Finally we validate the method on both synthetic data (\Cref{subsec:synthetic}), calcium recordings from the fruit fly head direction system (\Cref{subsec:drosophila}), and extracellular recordings from the mouse anterodorsal thalamic nucleus (\Cref{sec:peyrache}).

\subsection{Generative model \label{subsec:generative}}

We use $x_{ij}$ to denote the individual elements of a matrix $\b{X}$.
Let $\b{Y} \in \mathbb{R}^{N \times M}$ be the activity of $N$ neurons recorded in each of $M$ conditions.
Examples of ``conditions'' include time within a trial, stimulus identity, or motor output.
We assume that all neuronal responses collectively encode a shared, condition-specific latent variable $g_j \in \mathcal{M}$, where $\mathcal{M}$ is some manifold.
We further assume that each neuron $i$ is tuned to the latent state $g$ with a ``tuning curve'' $f_i(g)$, describing its average response conditioned on $g$.
Rather than assuming a specific parametric form for these tuning curves, we place a Gaussian process prior on $f_i(\cdot)$ to capture the heterogeneity widely observed in biological systems~\citep{churchland2007temporal,hardcastle2017multiplexed}.
The model is depicted in \Cref{fig:concept} and can be formally described as:
\begin{align}
    g_j          & \sim p^\mathcal{M}(g)                                           &  & \text{(prior over latents)}        \label{eq:latent_prior} \\
    f_i          & \sim \mathcal{GP}(0, k_i^\mathcal{M}(\cdot, \cdot)) &  & \text{(prior over tuning curves)} \label{eq:gp_prior}      \\
    y_{ij} | g_j & \sim \mathcal{N}(f_i(g_j), \sigma_i^2)              &  & \text{(noise model)} \label{eq:noise_model}
\end{align}
In \Cref{eq:latent_prior}, we use a uniform prior $p^\mathcal{M}(g)$ inversely proportional to the volume of the manifold for bounded manifolds (\Cref{subsec:priors}), and a Gaussian prior on Euclidean spaces to set a basic lengthscale.
In \Cref{eq:gp_prior}, $k_i^\mathcal{M}(\cdot,\cdot): \mathcal{M} \times \mathcal{M} \to \mathbb{R}$ is a covariance function defined on manifold $\mathcal{M}$ -- manifold-specific details are discussed in \Cref{subsec:tori_and_so3}.
In the special case where $\mathcal{M}$ is a Euclidean space, this model is equivalent to the standard Bayesian GPLVM \citep{Titsias2010a}.
While \Cref{eq:noise_model} assumes independent noise across neurons, noise correlations can also be introduced as in \citep{Wu2018} and Poisson noise as in \citep{Wu2017}.

This probabilistic model can be fitted by maximizing the log marginal likelihood
\begin{equation}\label{eq:trueL}
    \log{p(\b{Y})} = \log{ \int p(\b{Y}|\{ f_i \}, \{g_j\})\,p(\{f_i\}) \ p^\mathcal{M}(\{ g_j\})\ d\{f_i\} d\{g_j\}}.
\end{equation}
Following optimization, we can query both the posterior over latent states $p(\{g_j\}|\b{Y})$ and the posterior predictive distribution $p(\b{Y}^\star | \mathcal{G}^\star, \b{Y})$ at a set of query states $\mathcal{G}^\star$.
While it is possible to marginalise out $f_i$ when the states $\{ g_j \}$ are known, further marginalising out $\{ g_j \}$ is intractable and maximizing \Cref{eq:trueL} requires approximate inference.

\subsection{\label{subsec:inference}Learning and inference}

To maximize $\log{p(\b{Y})}$ in \Cref{eq:trueL}, we use variational inference as previously proposed for GPLVMs~\citep{Titsias2010a}.
The true posterior over the latent states, $p(\{g_j\}| {\bf Y})$, is approximated by a variational distribution $Q_\theta(\{ g_j \})$ with parameters $\theta$ that are optimized to minimize the KL divergence between $Q_\theta(\{ g_j \})$ and $p(\{ g_j \} | \b{Y})$.
This is equivalent to maximizing the evidence lower bound (ELBO) on the log marginal likelihood:
\begin{equation}\label{eq:elbo}
    \mathcal{L}(\theta) =
    H(Q_\theta) + \mathbb{E}_{Q_\theta}[\log p^\mathcal{M}(\{g_j\})]  + \mathbb{E}_{Q_\theta}[ \log{ p(\b{Y} | \{ g_j \}) } ].
\end{equation}
Here, $\mathbb{E}_{Q_{\theta}}[\cdot]$ indicates averaging over the variational distribution and $H(Q_\theta)$ is its entropy.
For simplicity, and because our model does not specify \emph{a priori} statistical dependencies between the individual elements of $\{ g_j \}$, we choose a variational distribution $Q_\theta$ that factorizes over conditions:
\begin{equation}
    \label{eq:q_factorization}
    Q_\theta(\{ g_j \}) = \prod_{j=1}^M q_{\theta_j}(g_j).
\end{equation}

In the Euclidean case, the entropy and expectation terms in \Cref{eq:elbo} can be calculated analytically for some kernels~\citep{Titsias2010a}, and otherwise using the reparameterization trick \citep{Kingma2014,Rezende2014}.
Briefly, the reparameterization trick involves first sampling from a fixed, easy-to-sample distribution (e.g.\ a normal distribution with zero mean and unit variance), and applying a series of differentiable transformations to obtain samples from $Q_\theta$.
We can then use these samples to estimate the entropy term and expectations in \Cref{eq:elbo}.

For non-Euclidean manifolds, inference in mGPLVMs poses two major problems.
Firstly, we can no longer calculate the ELBO analytically nor evaluate it using the standard reparameterization trick.
Secondly, evaluating the Gaussian process log marginal likelihood $\log p(\b{Y} | \{g_j\})$ exactly becomes computationally too expensive for large datasets.
We address these issues in the following.

\subsubsection{Reparameterizing distributions on Lie groups \label{subsubsec:relie}}

To estimate and optimize the ELBO in \Cref{eq:elbo} when $Q_\theta$ is defined on a non-Euclidean manifold, we use \citeauthor{Falorsi2019}'s ReLie framework, an extension of the standard reparameterization trick to variational distributions defined on Lie groups.

\paragraph{Sampling from $\b{Q}_\theta$}
Since we assume that $Q_{\theta}$ factorizes (\Cref{eq:q_factorization}), sampling from $Q_\theta$ is performed by independently sampling from each $q_{\theta_j}$.
We start from a differentiable base distribution $r_{\theta_j}(\b{x})$ in $\mathbb{R}^n$.
Note that $\mathbb{R}^n$ is isomorphic to the tangent space at the identity element of the group $G$, known as the Lie algebra.
We can thus define a `capitalized' exponential map $\Exp_G : \mathbb{R}^n \rightarrow G$, which maps elements of $\mathbb{R}^n$ to elements in $G$~(\citealp{Sola2018}; \Cref{appendix:lie}).
Importantly, $\Exp_G$ maps a distribution centered at zero in $\mathbb{R}^n$ to a distribution $\tilde{q}_{\theta_j}$ in the group centered at the identity element.
To obtain samples from a distribution $q_{\theta_j}$ centered at an arbitrary $g^{\mu}_j$ in the group, we can simply apply the group multiplication with $g^\mu_j$ to samples from $\tilde{q}_{\theta_j}$.
Therefore, obtaining a sample $g_j$ from $q_{\theta_j}$ involves the following steps: (i) sample from $r_{\theta_j}(\b{x})$,
(ii) apply $\Exp_G$ to obtain a sample $\tilde{g}_j$ from $\tilde{q}_{\theta_j}$,
and (iii) apply the group multiplication $g_j = g^\mu_j \tilde{g}_j$.

\paragraph{Estimating the entropy $\b{H}(\b{Q}_\theta)$}
Since $H(q_{\theta_j}) = H(\tilde{q}_{\theta_j})$~\citep{Falorsi2019}, we use $K$ independent Monte Carlo samples from $\tilde{Q}_{\theta}(\cdot)=\prod_{j=1}^M \tilde{q}_{\theta_j}(\cdot)$ to calculate
\begin{equation}
    H(Q_{\theta})
    \approx - \dfrac{1}{K}\sum_{k=1}^K \sum_{j=1}^M \log \tilde{q}_{\theta_j}( \tilde{g}_{jk}),
\end{equation}
where $\tilde{g}_{jk} = \Exp_G \b{x}_{jk}$ and $\{ \b{x}_{jk} \sim r_{\theta_j}(\b{x}) \}_{k=1}^K$.

\paragraph{Evaluating the density $\tilde{\b{q}}_\theta$}
To evaluate $\log \tilde{q}_{\theta_j}(\Exp_G{\b{x}_{jk}})$, we use the result from \citet{Falorsi2019} that
\begin{equation}\label{eq:density}
    \tilde{q}_\theta(\tilde{g}) = \sum_{\b{x} \in \mathbb{R}^n \: : \: \Exp_G{(\b{x})} = \tilde{g}}{ r_\theta(\b{x}) |\b{J}(\b{x})|^{-1} }
\end{equation}
where $\b{J}(\b{x})$ is the Jacobian of $\Exp_G$ at $\b{x}$.
Thus, $\tilde{q}_\theta(\tilde{g})$ is the sum of the Jacobian-weighted densities $r_\theta(\b{x})$ in $\mathbb{R}^n$ at \emph{all} those points that are mapped to $\tilde{g}$ through $\Exp_G{}$
This is an infinite but converging sum, and following \citet{Falorsi2019} we approximate it by its first few dominant terms (\Cref{sec:implementation}).

Note that $\Exp_G(\cdot)$ and the group multiplication by $g^\mu$ are both differentiable operations.
Therefore, as long as we choose a differentiable base distribution $r_\theta(\b{x})$, we can perform end-to-end optimization of the ELBO.
In this work we choose the reference distribution to be a multivariate normal $r_{\theta_j}(\b{x}) = \mathcal{N}(\b{x}; 0, \b{\Sigma}_j)$ for each $q_{\theta_j}$.
We variationally optimize both $\{ \b{\Sigma}_j \}$ and the mean parameters $\{ g^\mu_j \}$ for all $j$, and together these define the variational distribution.

\subsubsection{Sparse GP approximation \label{subsubsec:sparsegp}}

To efficiently evaluate the $\mathbb{E}_{Q_\theta}[ \log{ p(\b{Y} | \{ g_j \}) } ]$ term in the ELBO for large datasets, we use the variational sparse GP approximation~\citep{Titsias2009} which has previously been applied to Euclidean GPLVMs~\citep{Titsias2010a}.
Specifically, we introduce a set of $m$ inducing points $\mathcal{Z}_i$ for each neuron $i$, and use a lower bound on the GP log marginal likelihood:
\begin{align}\label{eq:sparsegp}
    \log{p(\b{y}_i|\{ g_j \})} & \geq \underbrace{-\dfrac{1}{2} \b{y}_i^T (\b{Q}_i + \sigma_i^2 \b{I})^{-1} \b{y}_i - \dfrac{1}{2}\log{|\b{Q}_i+\sigma_i^2 \b{I}|}  - \dfrac{1}{2 \sigma^2}\text{Tr}(\b{K}_i - \b{Q}_i) + \,\text{const.}}_{\displaystyle\log \tilde{p}(\b{y}_i|\{g_j \})} \\
    \text{with } \b{Q}_i       & = \b{K}_{\{g_j\}\mathcal{Z}_i} \b{K}_{\mathcal{Z}_i \mathcal{Z}_i}^{-1} \b{K}_{\mathcal{Z}_i \{ g_j\}}
\end{align}
where $\b{K}_{\mathcal{A}\mathcal{B}}$ denotes the Gram matrix associated with any two input sets $\mathcal{A}$ and $\mathcal{B}$.
Note that the latents $\{ g_j \}$ are shared across all neurons.
In this work we optimize the inducing points on $G$ directly, but they could equivalently be optimized in $\mathbb{R}^n$ and projected onto $G$ via $\Exp_G$.

Using the sparse GP framework, the cost of computing the GP likelihood reduces to $\mathcal{O}(M m^2)$ for each neuron and Monte Carlo sample.
This leads to an overall complexity of $\mathcal{O}(K N M m^2)$ for approximating $\mathbb{E}_{Q_\theta}[ \log{ p(\b{Y} | \{ g_j \}) }]$ with $K$ Monte Carlo samples, $N$ neurons, $M$ conditions and $m$ inducing points (see \Cref{sec:implementation} for further details on complexity and implementation).

\subsubsection{Optimization}

We are now equipped to optimize the ELBO defined in \Cref{eq:elbo} using Monte Carlo samples drawn from a variational distribution $Q_\theta$ defined on a Lie group $G$.
To train the model, we use Adam~\citep{kingma2014adam} to perform stochastic gradient descent on the following loss function:
\begin{equation}\label{eq:loss}
    \mathcal{L}(\theta) = \dfrac{1}{K} \sum_{k=1}^K \left[
    \sum_{j=1}^M { \left ( \log{p^{\mathcal{M}}( g_{jk} )}
    -  \log{ \tilde{q}_{\theta_j}( \tilde{g}_{jk} ) } \right ) }
    - \sum_i^N{ \log{\tilde{p}(\b{y}_i| \{ g_{jk} \} )}} \right]
\end{equation}
where a set of $K$ Monte-Carlo samples $\{ \tilde{g}_{jk} \}_{k=1}^K$ is drawn at each iteration from $\{ \tilde{q}_{\theta_j} \}$ as described in \Cref{subsubsec:relie}.
In \Cref{eq:loss}, $g_{jk} = g^\mu_j \tilde{g}_{jk}$, where $g^\mu_j$ is a group element that is optimized together with all other model parameters.
Finally, $\log{\tilde{p}(\b{y}_i| \{ g_j \})}$ is the lower bound defined in \Cref{eq:sparsegp} and $p^\mathcal{M}(g_{jk})$ is the prior described in \Cref{subsec:generative}.
The inner sums run over conditions $j$ and neurons $i$.

\subsubsection{\label{subsubsec:posterior_tuning}Posterior over tuning curves}

We approximate the posterior predictive distribution over tuning curves by sampling from the (approximate) posterior over latents.
Specifically, for a given neuron $i$ and a set of query states $\mathcal{G}^\star$, the posterior predictive over $\b{f}^\star_i$ is approximated by:
\begin{equation}
    \label{eq:posterior_tuning}
    p(\b{f}^\star_i | \b{Y}, \mathcal{G}^\star) = \frac1K \sum_{k=1}^K
    p(\b{f}^\star_i | \mathcal{G}^\star, \{ \mathcal{G}_k, \b{Y} \})
\end{equation}
where each $\mathcal{G}_k$ is a set of $M$ latent states (one for each condition in $\b{Y}$) independently drawn from the variational posterior $Q_\theta(\cdot)$.
In \Cref{eq:posterior_tuning}, each term in the sum is a standard Gaussian process posterior~\citep{williams2006gaussian}, which we approximate as described above (\Cref{subsubsec:sparsegp}; \Cref{sec:tuning_details}; \citealp{Titsias2009}).

\subsection{\label{subsec:tori_and_so3}Applying mGPLVM to tori, spheres and SO(3)}

At this stage, we have yet to define the manifold-specific GP kernels $k^\mathcal{M}$ described in \Cref{subsec:generative}.
These kernels ought to capture the topology of the latent space and express our prior assumptions that the neuronal tuning curves, defined on the manifold, have certain properties such as smoothness.
Here we take inspiration from the common squared exponential covariance function defined over Euclidean spaces and introduce analogous kernels on tori, spheres, and $SO(3)$.
This leads to the following general form:
\begin{equation}\label{eq:kernel}
    k^\mathcal{M}(g, g') = \alpha^2 \exp\left( - \dfrac{d_\mathcal{M}(g, g')}{2 \ell^2} \right) \qquad\qquad
    g, g' \in \mathcal{M}
\end{equation}
where $\alpha^2$ is a variance parameter, $\ell$ is a characteristic lengthscale, and $d_\mathcal{M}(g, g')$ is a manifold-specific distance function.
While squared geodesic distances might be intuitive choices for $d(\cdot,\cdot)$ in \Cref{eq:kernel}, they result in positive semi-definite (PSD) kernels only for Euclidean latent spaces~\citep{Jayasumana2015,feragen2015geodesic}.
Therefore, we build distance functions that automatically lead to valid covariance functions by observing that (i) dot product kernels are PSD, and (ii) the exponential of a PSD kernel is also PSD.
Specifically, we use the following manifold-specific dot product-based distances:
\begin{align}
    \label{eq:k_Rn}
    d_{R^n}(g, g')   & = ||g - g'||^2_2
                     & g \in \mathbb{R}^n
    \\
    \label{eq:k_Sn}
    d_{S^n}(g, g')   & = 2(1- g \cdot g')
                     &
    g \in \{\b{x} \in \mathbb{R}^{n+1} ;~ \| \b{x} \| = 1\}
    \\
    \label{eq:k_Tn}
    d_{T^n}(g, g')   & = \textstyle 2\sum_k{ ( 1 - g_k \cdot g'_k)  }      \qquad
                     & g \in \{ (g_1, \cdots, g_n) ; ~ \forall k: ~ g_k \in \mathbb{R}^2, \|g_k\|= 1 \}
    \\
    \label{eq:k_SO3}
    d_{SO(3)}(g, g') & = 4\left[1- \left ( g \cdot g' \right)^2 \right ]
                     & g \in \{\b{x} \in \mathbb{R}^{4} ;~ \| \b{x} \| = 1 \}
\end{align}
where we have slightly abused notation by directly using ``$g$'' to denote a convenient parameterisation of the group elements which we define on the right of each equation.
To build intuition, we note that the distance metric on the torus gives rise to a multivariate von Mises function; the distance metric on the sphere leads to an analogous von Mises Fisher function; and the distance metric on $SO(3)$ is $2(1-\cos \varphi_\text{rot})$ where $\varphi_\text{rot}$ is the angle of rotation required to transform $g$ into $g'$.
Notably, all these distance functions reduce to the Euclidean squared exponential kernel in the small angle limit.
Laplacian \citep{feragen2015geodesic} and Matérn \citep{borovitskiy2020matern} kernels have previously been proposed for modelling data on Riemannian manifolds, and these can also be incorporated in mGPLVM.

Finally, we provide expressions for the variational densities (\Cref{eq:density}) defined on tori, $S^3$ and $SO(3)$:
\begin{align}\label{eq:T_density}
    \tilde{q}_\theta( \Exp_{T^n}\b{x})
     & = \sum_{ \b{k} \in \mathbb{Z}^n}{ r_\theta( \b{x} + 2 \pi \b{k} ) },                                                 \\
    \label{eq:so3_density}
    \tilde{q}_\theta( \Exp_{SO(3)}\b{x})
     & =
    \sum_{k \in \mathbb{Z}}{ \left [ r_\theta(\b{x} + \pi k \hat{\b{x}}) \;
            \frac{2\|\b{x} + \pi k \hat{\b{x}}\|^2}{1 - \cos{\left ( 2\|\b{x} + \pi k \hat{\b{x}}\| \right ) }} \right ] }, \\
    \tilde{q}_\theta( \Exp_{S^3}\b{x})
     & =
    \sum_{k \in \mathbb{Z}}{ \left [ r_\theta(\b{x} + 2 \pi k \hat{\b{x}}) \;
            \frac{2\|\b{x} + 2 \pi k \hat{\b{x}}\|^2}{1 - \cos{\left ( 2\|\b{x} + 2 \pi k \hat{\b{x}}\| \right ) }} \right ] },
\end{align}
where $\hat{\b{x}} = \b{x}/\|\b{x}\|$.
Further details and the corresponding exponential maps are given in \Cref{subsec:manifold_densities}.
Since spheres that are not $S^1$ or $S^3$ are not Lie groups, ReLie does not provide a general framework for mGPLVM on these manifolds which we therefore treat separately in \Cref{subsec:spheres}.

%% file: results.tex
\section{Experiments and results}
\label{sec:results}

In this section, we start by demonstrating the ability of mGPLVM to correctly infer latent states and tuning curves in non-Euclidean spaces using synthetic data generated on $T^1$, $T^2$ and $SO(3)$.
We also verify that cross-validated model comparison correctly recovers the topology of the underlying latent space, suggesting that mGPLVM can be used for model selection given a set of candidate manifolds.
Finally, we apply mGPLVM to a biological dataset to show that it is robust to the noise and heterogeneity characteristic of experimental recordings.

\input{_ring.tex}

\subsection{Synthetic data}
\label{subsec:synthetic}

To generate synthetic data ${\bf Y}$, we specify a target manifold $\mathcal{M}$, draw a set of $M$ latent states $\{ g_j \}$ on $\mathcal{M}$, and assign a tuning curve to each neuron $i$ of the form
\begin{align}\label{eq:gfunc}
	f_i(g) &= a_i^2 \exp\left(
		- \dfrac{d^2_\text{geo}(g, g^\text{pref}_i)}{2 b_i^2}
		\right) + c_i,\\
	y_{ij} | g_j &\sim \mathcal{N}(f_i(g_j), \sigma_i^2)
\end{align}
with random parameters $a_i$, $b_i$ and $c_i$.
Thus, the activity of each neuron is a noisy bell-shaped function of the geodesic distance on $\mathcal{M}$ between the momentary latent state $g_j$ and the neuron's preferred state $g^\text{pref}_i$ (sampled uniformly).
While this choice of tuning curves is inspired by the common `Gaussian bump' model of neural tuning, we emphasize that the non-parametric prior over $f_i$ in mGPLVM can discover any smooth tuning curve on the manifold, not just Gaussian bumps.
For computational simplicity, here we constrain the mGPLVM parameters $\alpha_i$, $\ell_i$ and $\sigma_i$ to be identical across neurons.
Note that we can only recover the latent space up to symmetries which preserve pairwise distances.
In all figures, we have therefore aligned model predictions and ground truth for ease of visualization (\Cref{subsec:align}).

We first generated data on the ring ($T^1$, \Cref{fig:ring}, top left), letting the true latent state be a continuous random walk across conditions for ease of visualization.
We then fitted $T^1$-mGPLVM to the data and found that it correctly discovered the true latent states $g$ as well as the ground truth tuning curves (\Cref{fig:ring}, bottom right).
Reordering the neurons according to their preferred angles further exposed the population encoding of the angle (\Cref{fig:ring}, top right).

\input{_T2SO3.tex}

Next, we expanded the latent space to two dimensions with data now populating a 2-torus ($T^2$).
Despite the non-trivial topology of this space, $T^2$-mGPLVM provided accurate inference of both latent states (\Cref{fig:T2SO3}a) and tuning curves (\Cref{fig:T2SO3}b). 
To show that mGPLVM can be used to distinguish between candidate topologies, we compared $T^2$-mGPLVM to a standard Euclidean GPLVM in $\mathbb{R}^2$ on the basis of both cross-validated prediction errors and importance-weighted marginal likelihood estimates~\citep{burda2015importance}.
We simulated 10 different toroidal datasets; for each, we used half the conditions to fit the GP hyperparameters, and half the neurons to predict the latent states for the conditions not used to fit the GP parameters.
Finally, we used the inferred GP parameters and latent states to predict the activity of the held-out neurons at the held-out conditions.
As expected, the predictions of the toroidal model outperformed those of the standard Euclidean GPLVM which cannot capture the periodic boundary conditions of the torus (\Cref{fig:T2SO3}c).

Beyond toroidal spaces, $SO(3)$ is of particular interest for the study of neural systems encoding `yaw, pitch and roll' in a variety of 3D rotational contexts~\citep{shepard1971mental, Finkelstein2015, Wilson2018}.
We therefore fitted an $SO(3)$-mGPLVM to synthetic data generated on $SO(3)$ and found that it rendered a faithful representation of the latent space and outperformed a Euclidean GPLVM on predictions~(\Cref{fig:T2SO3}d-f).
Finally we show that mGPLVM can also be used to select between multiple non-Euclidean topologies.
We generated 10 datasets on each of $T^2$, $SO(3)$ and $S^3$ and compared cross-validated log likelihoods for $T^2$-, $SO(3)$- and $S^3$-mGPLVM, noting that $p(\mathcal{M} | \b{Y}) \propto p(\b{Y} | \mathcal{M})$ under a uniform prior over manifolds $\mathcal{M}$.
Here we found that the correct latent manifold was consistently the most likely for all 30 datasets (\Cref{fig:T2SO3}g).
In summary, these results show robust performance of mGPLVM across various manifolds of interest in neuroscience and beyond, as well as a quantitative advantage over Euclidean GPLVMs which ignore the underlying topology of the latent space.

\subsection{The \textit{Drosophila} head direction circuit}
\label{subsec:drosophila}

Finally we applied mGPLVM to an experimental dataset to show that it is robust to biological and measurement noise.
Here, we used calcium imaging data recorded from the ellipsoid body (EB) of \textit{Drosophila melanogaster}~\citep{turner2020neuroanatomical, turner2020kir}, where the so-called E-PG neurons have recently been shown to encode head direction~\citep{Seelig2015}.
The EB is divided into 16 `wedges', each containing 2-3 E-PG neurons that are not distinguishable on the basis of calcium imaging data, and we therefore treat each wedge as one `neuron'.
Due to the physical shape of the EB, neurons come `pre-ordered' since  their joint activity resembles a bump rotating on a ring (\Cref{fig:fly1d}a, analogous to \Cref{fig:ring}, ``ordered data'').
While the EB's apparent ring topology obviates the need for mGPLVM as an explorative tool for uncovering manifold representations, we emphasize that head direction circuits in higher organisms are not so obviously structured (\citet{Chaudhuri2019}; \Cref{sec:peyrache}) -- in fact, some brain areas such as the entorhinal cortex even embed concurrent representations of multiple spaces~\citep{hafting2005microstructure, constantinescu2016organizing}.

\input{_fly1d.tex}

We fitted the full mGPLVM with a separate GP for each neuron and found that $T^1$-mGPLVM performed better than $\mathbb{R}^1$-mGPLVM on both cross-validated prediction errors and log marginal likelihoods (\Cref{fig:fly1d}b).
The model recovered latent angles that faithfully captured the visible rotation of the activity bump around the EB, with larger uncertainty during periods where the neurons were less active (\Cref{fig:fly1d}a, orange).
When querying the posterior tuning curves from a fit in $\mathbb{R}^1$, these were found to suffer from spurious boundary conditions with inflated uncertainty at the edges of the latent representation -- regions where $\mathbb{R}^1$-mGPLVM effectively has less data than $T^1$-mGPLVM since $\mathbb{R}^1$ does not wrap around.
In comparison, the tuning curves were more uniform across angles in $T^1$ which correctly captures the continuity of the underlying manifold.
In \Cref{sec:peyrache}, we describe similar results with mGPLVM applied to a dataset from the mouse head-direction circuit with more heterogeneous neuronal tuning and no obvious anatomical organization~\citep{peyrache2015extracellular}.

%% file: _ring.tex
\begin{figure}[t!]
    \parbox{0.57\textwidth}{
        \begin{flushleft}
            \includegraphics[width=0.54\textwidth]{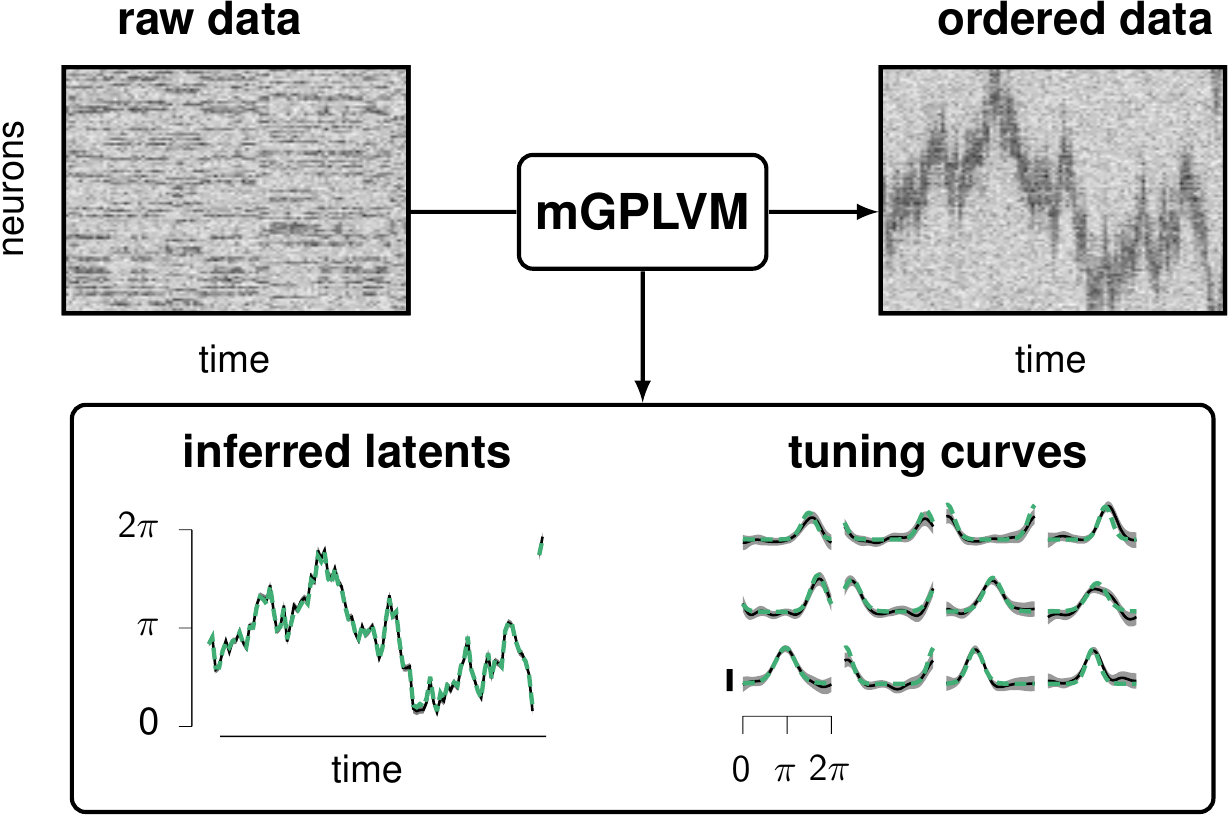}
        \end{flushleft}
    }
    \parbox{0.42\textwidth}{
        \caption{
            \label{fig:ring}
            {\bfseries Applying mGPLVM to synthetic data on the ring $T^1$.}
            {\bfseries Top left}: neural activity of $100$ neurons at $100$ different conditions (here, time bins). 
            {\bfseries Bottom}: timecourse of the latent states (left) and tuning curves for 12 representative neurons (right).
            Green: ground truth; Black: posterior mean; Grey shaded regions: $\pm$2 posterior s.t.d.
            {\bfseries Top right}: data replotted from the top left panel, with neurons reordered according to their preferred angles as determined by the inferred tuning curves.
        }
    }
    \vspace*{-1em}
\end{figure}

%% file: _T2SO3.tex
\begin{figure}[t!]
    \centering
    \includegraphics[width=0.99\textwidth]{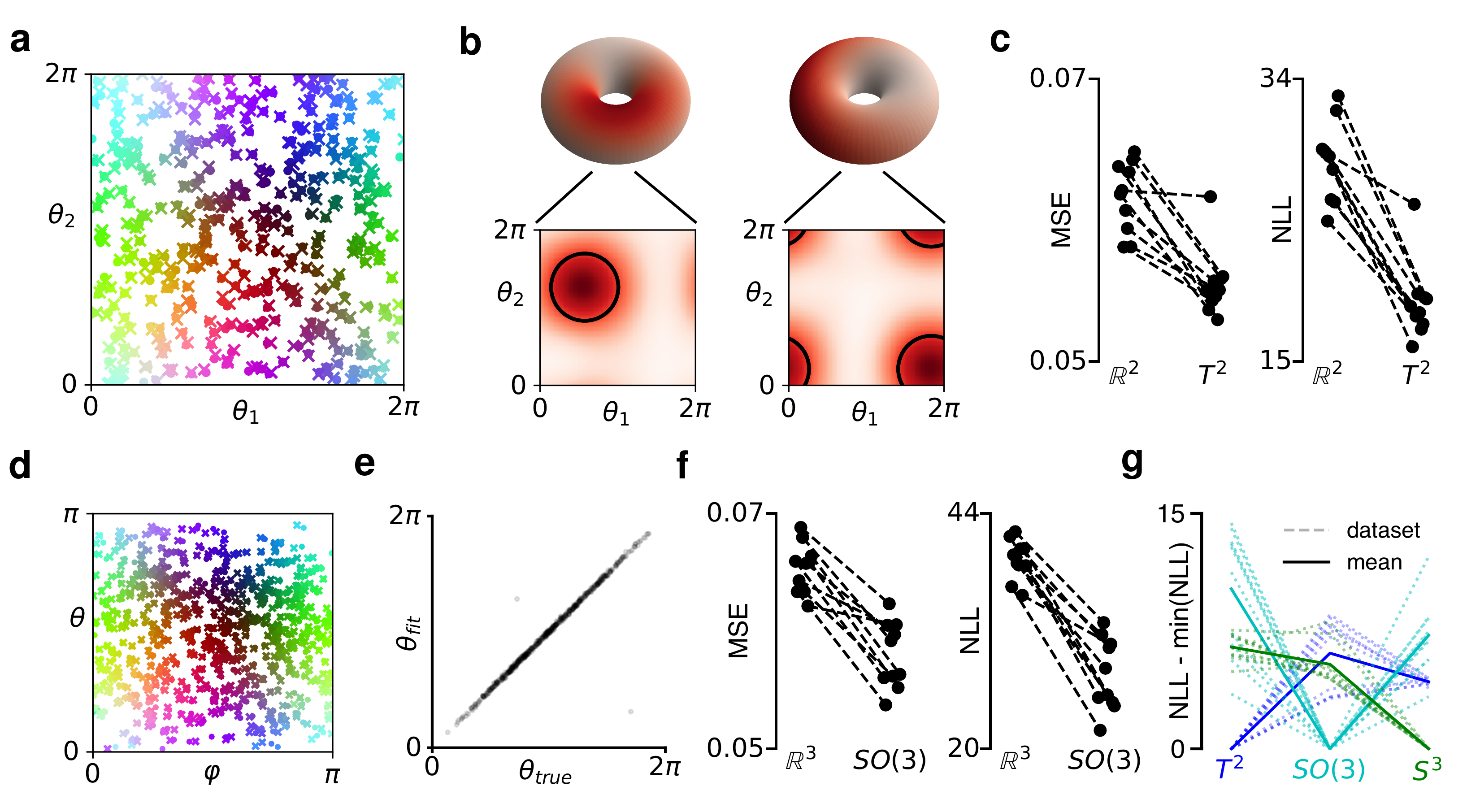}
    \caption{
        \label{fig:T2SO3}
        {\bfseries Validating mGPLVM on synthetic data.}
        {\bfseries (a-c)} Torus dataset.
        {\bfseries (a)} True latent states $\{ g_j \in T^2 \}$ (dots) and posterior latent means $\{  g^\mu_j \}$ (crosses).
        The color scheme is chosen to be smooth for the true latents.
        {\bfseries (b)} Posterior tuning curves for two example neurons.
Top: tuning curves on the tori.
Bottom: projections onto the periodic $[0; 2\pi]$ plane.
Black circles indicate locations and widths of the true tuning curves.
        {\bfseries (c)} Mean squared cross-validated prediction error (left) and negative log likelihood (right) when fitting $T^2$ and $\mathbb{R}^2$ to data generated on $T^2$.
        Dashed lines connect datapoints for the same synthetic dataset.
        {\bfseries (d-f)} $SO(3)$ dataset.
        {\bfseries (d)} 
	Axis of the rotation represented by the true latent states $\{ g_j \in SO(3) \}$ (dots) and the posterior latent means $\{ g^\mu_j \}$ (crosses) projected onto the $(\varphi, \, \theta)$-plane.
        {\bfseries (e)}
	Magnitude of the rotations represented by $\{ g_j \}$ and $\{ g^\mu_j \}$.
        {\bfseries (f)} Same as (c), now comparing $SO(3)$ to $\mathbb{R}^3$.
        {\bfseries (g)} Test log likelihood ratio for 10 synthetic datasets on $\color{blue} T^2$, $\color{cyan} SO(3)$, \& $\color{green} S^3$, with mGPLVM fitted on each manifold (x-axis).
        Solid lines indicate mean across datasets.
    }
    \vspace*{-1em}
\end{figure}

%% file: _fly1d.tex
\begin{figure}[t!]
    \parbox{0.65\textwidth}{
        \begin{flushleft}
            \includegraphics[width=0.62\textwidth]{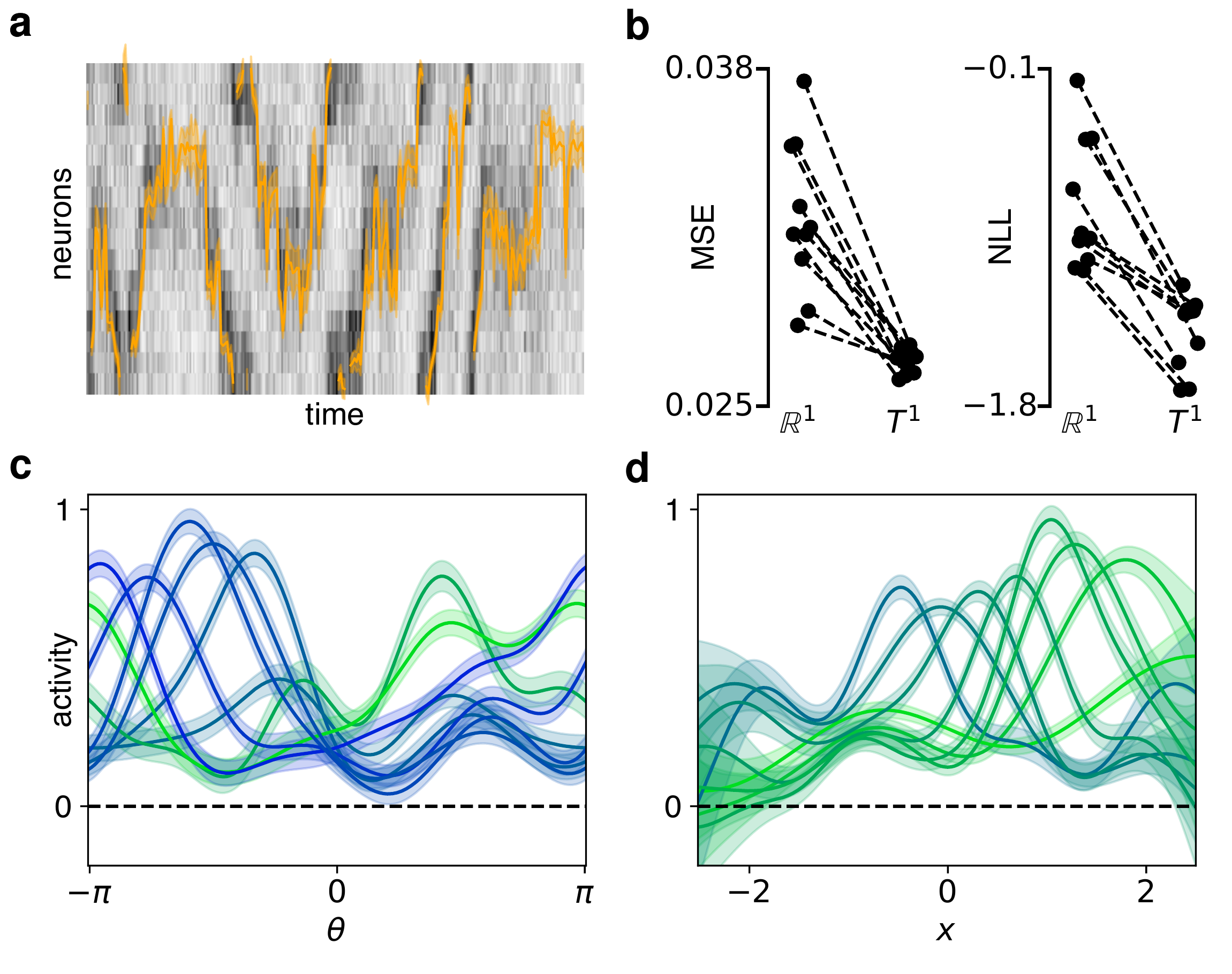}
        \end{flushleft}
    }
    \parbox{0.35\textwidth}{
    \caption{
    \label{fig:fly1d}
    {\bfseries The \textit{Drosophila} head direction circuit.}
    {\bfseries (a)}~
    Input data overlaid with the posterior variational distribution over latent states of a $T^1$-mGPLVM.
    {\bfseries (b)}~
    Mean cross-validated prediction error (left) and negative log likelihood (right) for models fitted on $T^1$ and $\mathbb{R}^1$.
    Each datapoint corresponds to a different partition of the timepoints into a training set and a test set.
        {\bfseries (c-d)}~
    Posterior tuning curves for eight example neurons in $T^1$ (c) and $\mathbb{R}^1$ (d).
    Color encodes the position of the maximum of each tuning curve. Shadings in (a,c,d) indicate $\pm 2$~s.t.d.
    }
    }
    \vspace*{-2em}
\end{figure}

%% file: discussion.tex
\section{Discussion and future work}
\label{sec:discussion}

\paragraph{Conclusion} 
We have presented an extension of the popular GPLVM model to incorporate non-Euclidean latent spaces.
This is achieved by combining a Bayesian GPLVM with recently developed methods for approximate inference in non-Euclidean spaces and a new family of manifold-specific kernels.
Inference is performed using variational sparse GPs for computational tractability with inducing points optimized directly on the manifold.
We demonstrated that mGPLVM correctly infers the latent states and GP parameters for synthetic data of various dimensions and topologies, and that cross-validated model comparisons can recover the correct topology of the space.
Finally, we showed how mGPLVM can be used to infer latent topologies and representations in biological circuits from calcium imaging data.
We expect mGPLVM to be particularly valuable to the neuroscience community because many quantities encoded in the brain naturally live in non-Euclidean spaces~\citep{Chaudhuri2019, Finkelstein2015, Wilson2018}.

\paragraph{Related work}
GP-based latent variable models with periodicity in the latent space have previously been used for motion capture, tracking and animation~\citep{urtasun2008topologically, elgammal2008tracking}. 
However, these approaches are not easily generalized to other non-Euclidean topologies and do not provide a tractable marginal likelihood which forms the basis of our Bayesian model comparisons.
Additionally, methods have been developed for analysing the geometry of the latent space of GPLVMs \citep{tosi2014metrics} and other latent variable models \citep{arvanitidis2017latent} after initially learning the models with a Euclidean latent.
These approaches confer a degree of interpretability to the learned latent space but do not explicitly incorporate priors and topological constraints on the manifold during learning.
Furthermore, GPs and GPLVMs with non-Euclidean outputs have been developed~\citep{Mallasto2018wrapped, navarro2017multivariate, mallasto2019probabilistic}.
These approaches are orthogonal to mGPLVM where the latent GP inputs, not outputs, live on a non-Euclidean manifold.
mGPLVM can potentially be combined with these approaches to model non-Euclidean observations, and to incorporate more expressive GP priors over the latent states than the independent prior we have used here.

Finally, several methods for inference in non-Euclidean spaces have been developed in the machine learning literature.
These have centered around methods based on VAEs~\citep{Davidson2018, wang2019riemannian,rey2019diffusion}, normalizing flows~\citep{Rezende2020}, and neural ODEs~\citep{lou2020neural,falorsi2020neural,mathieu2020riemannian}.
While non-Euclidean VAEs are useful for amortized inference, they constrain $f(g)$ more than a GP does and do not naturally allow expression of a prior over its smoothness.
Normalizing flows and neural ODEs can potentially be combined with mGPLVM to increase the expressiveness of the variational distributions \citep{Falorsi2019}.
This would allow us to model complex distributions over latents, such as the multimodal distributions that naturally arise in ambiguous environments with symmetries \citep{jacob2017independent}.

\paragraph{mGPLVM extensions}
Here, we have assumed statistical independence across latent states, but prior dependencies could be introduced to incorporate e.g.\ temporal smoothness by placing a GP prior on the latents as in GPFA~\citep{Yu2009}.
To capture more statistical structure in the latents, richer variational approximations of the posterior could be learned by using normalizing flows on the base distribution ($r_\theta$).
It would also be interesting to exploit automatic relevance determination (ARD, \citealp{Neal2012}) in mGPLVM to automatically select the latent manifold dimension.
We explored this approach by fitting a $T^2$-mGPLVM to the data from \Cref{fig:ring} with separate lengthscales for the two dimensions, where we found that $T^2$ shrunk to $T^1$, the true underlying manifold (\Cref{subsec:ard}).

Furthermore, the mGPLVM framework can be extended to direct products of manifolds, enabling the study of brain areas encoding non-Euclidean variables such as head direction jointly with global modulation parameters such as attention or velocity.
As an example, fitting a $(T^1 \times \mathbb{R}^1)$-mGPLVM to the \emph{Drosophila} data captures both the angular heading in the $T^1$ dimension as well as a variable correlated with global activity in the $\mathbb{R}^1$ dimension (\Cref{sec:lie_product}).

\paragraph{Future applications}
mGPLVM not only infers the most likely latent states but also estimates the associated uncertainty, which can be used as a proxy for the degree of momentary coherence expressed in neural representations. 
It would be interesting to compare such posterior uncertainties and tuning properties in animals across brain states.
For example, uncertainty estimates could be compared across sleep and wakefulness or environments with reliable and noisy spatial cues.

In the motor domain, mGPLVM can help elucidate the neural encoding of motor plans for movements naturally specified in rotational spaces.
Examples include 3-dimensional head rotations represented in the rodent superior colliculus~\citep{Wilson2018,masullo2019genetically} as well as analogous circuits in primates.
Finally, it will be interesting to apply mGPLVM to artificial agents trained on tasks that require them to form internal representations of non-Euclidean environmental variables~\citep{Banino2018}.
Our framework could be used to dissect such representations, adding to a growing toolbox for the analysis of artificial neural networks~\citep{sussillo2013opening}.

%% file: acknowledgements.tex
\subsection*{Acknowledgements}
We thank Daniel Turner-Evans and Vivek Jayaraman for sharing their experimental data.
K.T.J. was funded by a Gates Cambridge scholarship; T-C.K. by a Trinity-Henry Barlow scholarship and a scholarship from the Ministry of Education, ROC Taiwan; and M.T. by the Medical Research Council (MC\_UP\_12012) and an ERC Starting Grant (STG 677029).
We are grateful for helpful comments on the manuscript by Robert Pinsler, Marine Schimel, David Liu, and others in the CBL.

%% file: appendix.tex
\normalsize

\section{The mouse head direction circuit}
\label{sec:peyrache}

\input{_mouse.tex}

To highlight the importance of unsupervised non-Euclidean learning methods in neuroscience and to illustrate the interpretability of the learned GP parameters, we consider a dataset from \citetAPP{peyrache2015extracellularAPP} recorded from the mouse anterodorsal thalamic nucleus (ADn; \Cref{fig:mouse}a).
This data has also been analyzed in \citetAPP{peyrache2015internallyAPP}, \citetAPP{chaudhuri2019intrinsicAPP} and \citetAPP{rubin2019revealingAPP}.
We consider the same example session shown in Figure 2 of \citetAPP{chaudhuri2019intrinsicAPP} (Mouse 28, session 140313) and bin spike counts in 500~ms time bins for analysis with mGPLVM.
When comparing cross-validated log likelihoods for $T^1$- and $\mathbb{R}^1$-mGPLVM fitted to the data, $T^1$ consistently outperformed $\mathbb{R}^1$ with a log likelihood ratio of $127 \pm 30$ (mean $\pm$ sem) across 10 partitions of the data.

Fitting $T^1$-mGPLVM to the binned spike data, we found that the inferred latent state was highly correlated with the true head direction (\Cref{fig:mouse}b).
However, in contrast to the data considered in \Cref{subsec:synthetic} and \Cref{subsec:drosophila}, this mouse dataset contains neurons with more heterogeneous baseline activities and tuning properties.
This is reflected in the learned GP parameters which converge to small kernel length scales for neurons that contribute to the heading representation (\Cref{fig:mouse}c, `tuned') and large length scales for those that do not (\Cref{fig:mouse}c, `not tuned').
Finally, since mGPLVM does not require knowledge of behaviour, we also fitted mGPLVM to data recorded from the same neurons during a period of rapid eye movement (REM) sleep.
Here we found that the representation of subconscious heading during REM sleep was similar to the representation of heading when the animal was awake after matching the offset between the two sets of tuning curves (\Cref{fig:mouse}d), similar to results by \citetAPP{peyrache2015internallyAPP}.
However, their analyses relied on recordings from two separate brain regions to align the activity from neurons in ADn to a subconscious head direction decoded from the postsubiculum and vice versa.
In contrast, mGPLVM allows for fully unsupervised Bayesian analyses across both wake and sleep using recordings from a single brain area.

\section{Priors on manifolds}\label{subsec:priors}

For all manifolds, we use priors that factorize over conditions, $p^{\mathcal{M}}(\{g_j\}) = \prod_j{p^{\mathcal{M}}(g_j)}$. 
As described in \Cref{subsec:generative}, we use a Gaussian prior 
$p^{R^n}(g) = \mathcal{N}(g; 0, {\bf I}_n)$ over latent states in $\mathbb{R}^n$, and uniform priors for the spheres, tori, and $SO(3)$.
These uniform priors have a density which is the inverse volume of the manifold: 
\begin{align}
	p^{S^n}(g) & = \left [ \dfrac{2 \pi^{\tfrac{n+1}{2}}}{\Gamma(\tfrac{n+1}{2})} \right ]^{-1}\\
	p^{T^n}(g) & = [2 \pi]^{-n}\\
    p^{SO(3)}(g) & = \left [ \dfrac{2 \pi^{\tfrac{4}{2}}}{2 \Gamma(\tfrac{4}{2})} \right ]^{-1}.
\end{align}
Note that the volume of $S^n$ is the surface area of the $n$-sphere, and the volume of $SO(3)$ is half the volume of $S^3$.

\section{Lie groups and their exponential maps \label{appendix:lie}}

For simplicity of exposition, we have skimmed over the details of how the `capitalized' Exponential map $\Exp_G : \mathbb{R}^n \rightarrow G$ is defined in \Cref{subsubsec:relie}, particularly in relation to the group's Lie algebra $\mathfrak{g}$. 
Here we make this connection more explicit.
As described in the main text, the Lie algebra $\mathfrak{g}$ of a group $G$ is a vector space tangent to $G$ at its identity element.
The exponential map $\exp_G : \mathfrak{g} \rightarrow G$ maps elements from the Lie algebra to the group, and is conceptually distinct from the ``capitalised'' Exponential map defined in \Cref{subsubsec:relie} which maps from $\mathbb{R}^n$ to $G$.
However, because the Lie algebra is isomorphic to $\mathbb{R}^n$, we have found it convenient in both our exposition and our implementation to work directly with the pair $(\mathbb{R}^n, \Exp_G)$, instead of $(\mathfrak{g}, \exp_G)$.
To expand on the connection between the two, note that we can define as in \citetAPP{Sola2018APP} the isomorphism $\text{Hat} : \mathbb{R}^n \rightarrow \mathfrak{g}$, which maps every element in $\mathbb{R}^n$ to a distinct element in the Lie algebra $\mathfrak{g}$.
Therefore, $\Exp_G : \mathbb{R}^n \rightarrow G$ is in fact the composition $\exp_G \circ \text{Hat}$.

\subsection*{Manifold-specific parameterizations}
\label{subsec:manifold_densities}

Here we provide some further justification for the forms of $\tilde{q}_\theta(\tilde{g})$ provided in \Cref{eq:T_density,eq:so3_density} as well as the exponential maps which are used to derive these densities and are needed for optimization in \Cref{eq:loss}.
For both $T^n$ and $SO(3)$, we use \Cref{eq:density} from \citetAPP{Falorsi2019APP}, which we repeat here for reference:
\begin{equation}\label{eq:density_app}
    \tilde{q}_\theta(\tilde{g}) = \sum_{\b{x} \in \mathbb{R}^n \: : \: \Exp_G{(\b{x})} = \tilde{g}}{ r_\theta(\b{x}) |\b{J}(\b{x})|^{-1} }.
\end{equation}
In what follows, we will use $\b{g}$ to indicate a vector representation of group element $g$ to avoid conflicts of notation.

Note that the expressions in this section largely follow \citetAPP{Falorsi2019APP}, but we re-write them in a different basis for ease of computational implementation.

\subsection{$T^n$}
\label{appendix:subsec:param-n-torus}

The $n$-Torus $T^n$ is the direct product of $n$ circles, such that we can parameterize members of this group as $\b{g} \in \mathbb{R}^n$ whose elements are all angles between $0$ and $2\pi$.
Note that this is equivalent to the parameterization in \Cref{eq:k_Tn} except that here we denote an element on the circle by its angle, while in \Cref{eq:k_Tn} we denote it by a unit $2$-vector for notational consistency with the other kernels.
Because 1-dimensional rotations are commutative, the parameterization of the torus as a list of angles allows us to perform group operations by simple addition modulo $2\pi$.
We therefore slightly abuse notation and write the exponential map $\Exp_{T^n} : \mathbb{R}^n \rightarrow T^n$ as an element-wise modulo operation:
\begin{equation}\label{eq:expT}
    \Exp_{T^n}\b{x} = \b{x} \; \text{mod} \; 2 \pi.
\end{equation}
\Cref{eq:expT} has inverse Jacobian $|\b{J}(x)|^{-1} = 1$.
Moreover, since $\Exp_{T^n}(\b{x}) = \Exp_{T^n}(\b{x} + 2\pi \b{k})$ for any integer vector $\b{k} \in \mathbb{Z}^n$, the change-of-variable formula in \Cref{eq:density_app} yields the following density on $T^n$:
\begin{equation}\label{eq:T_density_app}
    \tilde{q}_\theta( \Exp_{T^n}\b{x}) = \sum_{ \b{k} \in \mathbb{Z}^n}{ r_\theta( \b{x} + 2 \pi \b{k} ) }.
\end{equation}
For ease of implementation it is also convenient to rewrite the kernel distance function \Cref{eq:k_Tn} as 
\begin{equation}\label{eq:k_Tn_app}
    d_{T^n}(\b{g}, \b{g}') = 2 \cdot \b{1}_n \cdot (1 - \cos(\b{g}-\b{g}'))
\end{equation}
where $\b{1}_n$ is the n-vector full of ones, and $\cos(\cdot)$ is applied element-wise to $\b{g}-\b{g'}$.

\subsection{$SO(3)$}
\label{appendix:subsec:param-so3}

We use quaternions $\b{g} \in \mathbb{R}^4$ to represent elements $g \in SO(3)$ as indicated in \Cref{eq:k_SO3}.
For a rotation of $\phi$ radians around axis $\b{u} \in \mathbb{R}^3$ with $\| \b{u} \|=1$,
\begin{equation}
    \b{g} = \left(\cos\frac\phi2, \b{u} \sin\frac\phi2 \right)
    \in \mathbb{R}^4.
\end{equation}
The exponential map $\Exp_{SO(3)} : \mathbb{R}^3 \rightarrow SO(3)$ is
\begin{equation}\label{eq:so3exp}
    \Exp_{SO(3)}\b{x} = ( \cos{\|\b{x}\|}, \hat{\b{x}} \sin{\|\b{x}\|}),
\end{equation}
where $\hat{\b{x}} = \b{x}/\|\b{x}\|$ and $\phi = 2 \|\b{x}\|$ is the angle of rotation.
This gives rise to an inverse Jacobian 
\begin{equation}
|\b{J}(\b{x})|^{-1} = \phi^2/(2(1 - \cos{\phi})).
\end{equation}
Using \Cref{eq:density_app} we get the density on the group
\begin{equation}\label{eq:so3_density_app}
    \tilde{q}_\theta( \Exp_{SO(3)}\b{x}) =
    \sum_{k \in \mathbb{Z}}{ \left [ r_\theta(\b{x} + \pi k \hat{\b{x}}) \;
    \frac{2\|\b{x} + \pi k \hat{\b{x}}\|^2}{1 - \cos{\left ( 2\|\b{x} + \pi k \hat{\b{x}}\| \right ) }} \right ] },
\end{equation}
where the sum over $k$ stems from the fact that a rotation of $\phi + 2 k \pi$ around axis $\hat{\b{x}}$ is equivalent to a rotation of $\phi$ around the same axis.

\section{mGPLVM on $S^n$ \label{subsec:spheres}}

In this section, we discuss how to fit mGPLVMs on spheres.
We first consider spheres which are also Lie groups, and then discuss a general framework for all $n$-spheres.

\subsection{ $S^{1, 3}$ }

We begin by noting that $S^{n}$ is not a Lie group unless $n=1$ or $n=3$, thus we can only apply the ReLie framework to $S^1$ and $S^3$.
$S^1$ is equivalent to $T^1$ and is most easily treated using the torus formalism above.
For $S^3$, we note that $SO(3)$ is simply $S^3$ with double coverage. This is because quaternions $\b{g}$ and $-\b{g}$ represent the same element of $SO(3)$ while they correspond to distinct elements of $S^3$.
The Jacobian and exponential maps of $S^3$ are therefore identical to those of $SO(3)$.
The expression for the density on $S^3$ also mirrors \Cref{eq:so3_density_app} except that the sum is over $\b{x} + 2 \pi k \hat{\b{x}}$ instead of $\b{x} + \pi k \hat{\b{x}}$:
\begin{equation}\label{eq:s3_density_app}
    \tilde{q}_\theta( \Exp_{S^3}\b{x}) =
    \sum_{k \in \mathbb{Z}}{ \left [ r_\theta(\b{x} + 2 \pi k \hat{\b{x}}) \;
    \frac{2\|\b{x} + 2 \pi k \hat{\b{x}}\|^2}{1 - \cos{\left ( 2\|\b{x} + 2 \pi k \hat{\b{x}}\| \right ) }} \right ] }.
\end{equation}
We demonstrate $S^3$-mGPLVM on synthetic data from $S^3$ in \Cref{fig:sphere} (bottom).

\input{_sphere.tex}

\subsection{ $S^{n \notin \{1, 3 \}}$ }

The ReLie framework does not directly apply to distributions defined on non-Lie groups.
Nevertheless, we can still apply mGPLVM to an $n$-sphere embedded in $\mathbb{R}^{n+1}$ by taking each latent variational distribution $q_{\theta_j} $ to be a von Mises-Fisher distribution (VMF), whose entropy is known analytically.
Parameterizing group element $g \in S^n$ by a unit-norm vector $\b{g} \in \mathbb{R}^{n+1}$, $\| \b{g} \| = 1$, this density is given by:
\begin{equation}
    q_\theta(\b{g}; \b{g}^\mu, \kappa) = 
    \frac{\kappa^{n/2-1}}{(2\pi)^{n/2} I_{n/2-1}(\kappa)}
    \exp(\kappa \, {\b{g}^\mu} \cdot \b{g})
\end{equation}
where $\cdot$ denotes the dot product. 
Here, $I_v$ is the modified Bessel function of the first kind at order $v$, $\b{g}^\mu$ is the mean direction of the distribution on the hypersphere, and $\kappa \geq 0$ is a concentration parameter -- the larger $\kappa$, the more concentrated the distribution around $\b{g}^\mu$.

Using a VMF distribution as the latent distribution, we can easily evaluate the ELBO in \Cref{eq:elbo} because (i) there are well-known algorithms for sampling from the distribution using rejection-sampling \citepAPP{Ulrich1984APP} and (ii) both the entropy term $H(q_\theta)$ and its gradient can be derived analytically \citepAPP{Davidson2018APP}.
For details of how to differentiate through rejection sampling, please refer to \citetAPP{Naesseth2016APP} and \citetAPP{Davidson2018APP}.

In the following, we provide details for applying mGPLVM to $S^2$ for which we do not need to use rejection sampling and instead use inverse transform sampling  \citepAPP{Jakob2012APP}.
For $S^2$, the VMF distribution simplifies to~\citepAPP{Straub2017APP} 
\begin{equation}
    q_\theta(\b{g}; \b{g}^\mu, \kappa) = 
    \frac{\kappa}{2\pi(\exp (\kappa) - \exp (-\kappa))}
    \exp(\kappa \, {\b{g}^\mu} \cdot \b{g}),
\end{equation}
and its entropy is 
\begin{align}
    H(q_\theta) &= - \int_{S^2} q_\theta(\b{g}; \b{g}^\mu, \kappa) \log q_\theta(\b{g}; \b{g}^\mu, \kappa) d\b{g}\\
      &= - \log \bigg ( \frac{\kappa}{4\pi \sinh \kappa} \bigg ) - \frac{\kappa}{\tanh \kappa} + 1.
\end{align}
These equations allow us to apply mGPLVM to $S^2$ by optimizing the ELBO as described in the main text; this is illustrated for synthetic data on $S^2$ in \Cref{fig:sphere} (top).

\section{Posterior over tuning curves}
\label{sec:tuning_details}

We can derive the posterior over tuning curves 
in \Cref{eq:posterior_tuning} as follows:
\begin{align}
    \label{eq:posterior_tuning_APP}
p(\b{f}^\star_i | \b{Y}, \mathcal{G}^\star) 
&= 
\int p(\b{f}^\star_i, \mathcal{G}| \mathcal{G}^\star, \b{Y}) ~d \mathcal{G} \\
&=\int p(\b{f}^\star_i| \mathcal{G}^\star, \{\mathcal{G}, \b{Y}\}) 
p(\mathcal{G}  | \b{Y})
~d \mathcal{G} \\
&\approx \int p(\b{f}^\star_i| \mathcal{G}^\star, \{\mathcal{G}, \b{Y}\}) 
Q_\theta( \mathcal{G} )
~d \mathcal{G} \\
&\approx\frac1K \sum_{k=1}^K 
    p(\b{f}^\star_i | \mathcal{G}^\star, \{ \mathcal{G}_k, \b{Y} \})
\end{align}
where each $\mathcal{G}_k$ is a set of $M$ latents (one for each of the $M$ conditions in the data $\b{Y}$) sampled from the variational posterior $Q_\theta(\mathcal{G})$.
The standard deviation around the mean tuning curves in all figures are estimated from $1000$ independent samples from this posterior, with each draw involving the following two steps: (i) draw a sample $\mathcal{G}_k$ from $Q_\theta$ and (ii) conditioned on this sample, draw from the predictive distribution 
$p(\b{f}^\star_i | \mathcal{G}^\star, \{ \mathcal{G}_k, \b{Y} \})$. 
Together, these two steps correspond to a single draw from the posterior.
Note that we make a variational sparse GP approximation (\Cref{subsubsec:sparsegp}) and therefore approximate the predictive distribution $p(\b{f}^\star_i | \mathcal{G}^\star, \{ \mathcal{G}_k, \b{Y} \})$ as described in \citetAPP{Titsias2009APP}.

\section{Alignment for visualization}\label{subsec:align}

The mGPLVM solutions for non-Euclidean spaces are degenerate because the ELBO depends on the sampled latents through (i) their uniform prior density, (ii) their entropy, and (iii) the GP marginal likelihood, and all three quantities are invariant to transformations that preserve pairwise distances.
For example, the application of a common group element $g$ to \emph{all} the variational means leaves pairwise distances unaffected and therefore does not affect the ELBO.
Additionally, pairwise distances are invariant to reflections along any axis of the coordinate system we have chosen to represent each group. 
Therefore, to plot comparisons between true and fitted latents, we use numerical optimization to find a single distance-preserving transformation that minimizes the average geodesic distance between the variational means $\{  g^\mu_j \}$ and the true latents $\{ g_j\}$.

For the $n$-dimensional torus (\Cref{fig:ring,fig:T2SO3}) which we parameterize as 
$$\b{g} \in \{(g_1, \cdots, g_n); \forall k: g_k \in [0, 2\pi]\},$$
the distance metric depends on $\cos(g_k - g_k')$ and is invariant to any translation and reflection of all latents along each dimension
$$ g_k \to  (\alpha_k g_k + \beta_k) \mod 2\pi$$
where $\alpha_k \in \{1, -1\}$ and $\beta_k \in [0, 2\pi]$.
We optimize discretely over the $\{\alpha_k \}$ by trying every possible combination, and continuously over $\beta_k$ for each combination of $\{ \alpha_k \}$.

In the case of $S^2$, $S^3$ and $SO(3)$ (\Cref{fig:T2SO3,fig:sphere}), the distance metrics are invariant to unitary transformations $\b{g} \to \b{R} \b{g}$ where $\b{R} \b{R}^T = \b{R}^T \b{R} = \b{I}$ for the parameterizations used in this work.
For visualization of these groups, we align the inferred latents with the true latents by optimizing over $\b{R}$ on the manifold of orthogonal matrices.

\section{Automatic relevance determination}\label{subsec:ard}

\input{_ard.tex}

As we mention in \Cref{sec:discussion}, it is possible to exploit automatic relevance determination (ARD) for automatic selection of the dimensionality of groups with additive distance metrics such as the $T^n$-distance in \Cref{eq:k_Tn_app}.
While we have not investigated this in detail, we illustrate the idea here on a simple example.
We consider the same synthetic data as in \Cref{fig:ring} and fit a $T^2$-mGPLVM with a kernel on $T^2$ that has separate lengthscales $\ell_1$ and $\ell_2$ for each dimension:
\begin{equation}
k_{T^2_\text{ARD}}(\b{g}, \b{g}') = \alpha^2 \exp\left(\dfrac{\cos(g_1 - g_1')-1}{\ell_1^2}\right)
\,
\exp\left(\dfrac{\cos(g_2 - g_2')-1}{\ell_2^2} \right).
\end{equation}
Additionally, we assume the variational distribution to factorize across latent dimensions:
\begin{equation}
q_{\theta_j}(\cdot) = q_{\theta^1_j}(\cdot)\, q_{\theta^2_j}(\cdot),
\end{equation}
such that their entropies add up to the total entropy:
\begin{equation}
H(q_{\theta_j}) = H(q_{\theta^1_j}) + H(q_{\theta^2_j}).
\end{equation}
This corresponds to assuming that each variational covariance matrix ${\bf \Sigma}_j$ (\Cref{subsubsec:relie}) is diagonal.

When fitting this model, we find that one length parameter goes to large values while the other remains on the order of the size of the space (\Cref{fig:ard}a; note that $d_{T^1} \in [0, 4] $).
This indicates that neurons are only tuned to one of the two torus dimensions.  
Additionally, posterior variances become very large in the non-contributing dimension, i.e.\ the data does not contain the other angular dimension (\Cref{fig:ard}b).
This further indicates that the model has effectively shrunk from a 2-torus to a single circle.
We note that the entropy of the factor in the variational posterior that corresponds to the discarded dimension becomes $\log2\pi$ as the variance goes to infinity in this direction.
This exactly offsets the increased complexity penalty of the prior for $T^2$ compared to $T^1$, such that the two models have the same ELBO.
The model thus reduces to a $T^1$ model, demonstrating how ARD can be exploited to automatically infer the dimensionality of the latent space.

\section{Direct products of Lie groups}
\label{sec:lie_product}

Here, we elaborate slightly on the extension of mGPLVM to direct products of Lie groups, briefly mentioned in the discussion (\Cref{sec:discussion}).
Assuming additive distance metrics and factorizable variational distributions, direct product kernels become multiplicative and entropies become additive -- very much as in our illustration of ARD in \Cref{subsec:ard}.
That is, for a group product $\mathcal{M} = \mathcal{M}_1 \times \ldots \times \mathcal{M}_L$, we can write
\begin{align}
	k^\mathcal{M}(g, g') &= \prod_l{ k^{\mathcal{M}_l} (g, g') },\\
	H(q^\mathcal{M}_{\theta_j}) &= \sum_l{  H(q^{\mathcal{M}_l}_{\theta_j})  }.
\end{align}
As a simple example, we consider a $(T^1 \times \mathbb{R}^1)$-mGPLVM which we fit to the \emph{Drosophila} data from \Cref{subsec:drosophila}.
Here we find that the $T^1$ dimension of the group product, which we denote by $\theta^{(T^1 \times \mathbb{R}^1)}$, captures the angular component of the data since it is very strongly correlated with the latent state $\theta^{T^1}$ inferred by the simpler $T^1$-mGPLVM (\Cref{fig:fly2d}a).
It is somewhat harder to predict what features of the data will be captured by the $\mathbb{R}^1$ dimension $x^{(T^1 \times \mathbb{R}^1)}$ of the $(T^1 \times \mathbb{R}^1)$-mGPLVM, but we hypothesize that it might capture a global temporal modulation of the neural activity.
We therefore plot the mean instantaneous activity $\bar{y}$ across neurons against $x^{(T^1 \times \mathbb{R}^1)}$ and find that these quantities are indeed positively correlated (\Cref{fig:fly2d}b).
This exemplifies how an mGPLVM on a direct product of groups can capture qualitatively different components of the data by combining representations with different topologies.

This direct product model is very closely related to the ARD model in \Cref{subsec:ard}, and the two can also be combined in a direct product of ARD kernels.
For example, we can imagine constructing a $(T^n \times \mathbb{R}^n)$ direct product ARD kernel which automatically selects the appropriate number of both periodic and scalar dimensions that best, and most parsimoniously, explains the data.

\input{_fly2d.tex}

\section{Implementation}
\label{sec:implementation}

\paragraph{Scaling}
As mentioned in \Cref{subsubsec:sparsegp}, approximating the GP likelihood term $\mathbb{E}_{Q_\theta}[ \log{ p(\b{Y} | \{ g_j \}) }]$ in the mGPLVM ELBO scales as $\mathcal{O}(m^2 M N K)$ with $m$ inducing points, $M$ latent states, $N$ neurons, and $K$ Monte Carlo samples.
Estimating the entropy term is $\mathcal{O}(M K d)$ for a $d$-dimensional Euclidean latent space, $\mathcal{O}(M K (2k_{max} +1)^d)$ for a d-dimensional torus, and $\mathcal{O}(M K (2k_{max} +1))$ for $SO(3)$ and $S^3$, where $k_{max}$ is the maximum value of $k$ used in \Cref{eq:density}.
For all manifolds considered in this work, we can compute a closed-form $\Exp(\cdot)$ while for general matrix Lie groups, approximating $\Exp$ as a power series is $\mathcal{O}(d^3)$ \citepAPP{Falorsi2019APP}, further increasing the complexity of mGPLVM for such groups.

For our manifolds of interest, computing the likelihood term tends to be the main computational bottleneck, although the entropy term can become prohibitive for high-dimensional periodic latents \citepAPP{Rezende2020APP}. 
When computing $\mathbb{E}_{Q_\theta}[ \log{ p(\b{Y} | \{ g_j \}) }]$, most of the complexity is due to inverting $N K$ matrices of size $(M m^2) \times (M m^2)$, which can be performed in parallel for each Monte Carlo sample and neuron.
Using PyTorch for parallelization across neurons and MC samples, we can train $T^1$-mGPLVM with $N=300$ and $M=1000$ in $\sim\!100$ seconds on an NVIDIA GeForce RTX 2080 GPU with 8GB RAM.

\paragraph{Initialization}
For all simulations, we initialized the system with variational means at the identity element of the manifold, but with large variational variances to reflect the lack of prior information about the true latent states.
Inducing points were initialized according to the prior on each manifold (\Cref{eq:latent_prior}).
To avoid variational distributions collapsing to the uniform distribution early during learning, we ran a preliminary `warm up' optimization phase during which some of the parameters were held fixed.
Specifically, we fixed the variational covariance matrices as well as the kernel variance parameters ($\alpha$ in \Cref{eq:kernel}), and prioritized a better data fit by setting the entropy term to zero in \Cref{eq:elbo}.
Learning proceeded as normal thereafter. 

\paragraph{Entropy approximation}
When evaluating \Cref{eq:density}, we used values of $k_{max}=3$ for the tori and $S^3$ as in \citetAPP{Falorsi2019APP} and $k_{max} = 5$ for $SO(3)$ since the sum takes steps of $\pi$ instead of $2\pi$.
In theory, the finite $k_{max}$ can lead to an overestimation of the ELBO for large variational uncertainties, as $\tilde{q}$ is systematically underestimated, leading to overestimation of the entropy.
To mitigate this, we capped the approximate entropy for non-Euclidean manifolds at the maximum entropy corresponding to a uniform distribution on the manifold.

\paragraph{Code}
A python package implementing mGPLVM can be found at \url{https://github.com/tachukao/mgplvm-pytorch}, including instructions for downloading the datasets used in the paper and running several example calculations.

%% file: _mouse.tex
\begin{figure}[h!]
    \vspace*{-2em}
    \parbox{0.65\textwidth}{
        \begin{flushleft}
            \includegraphics[width=0.62\textwidth]{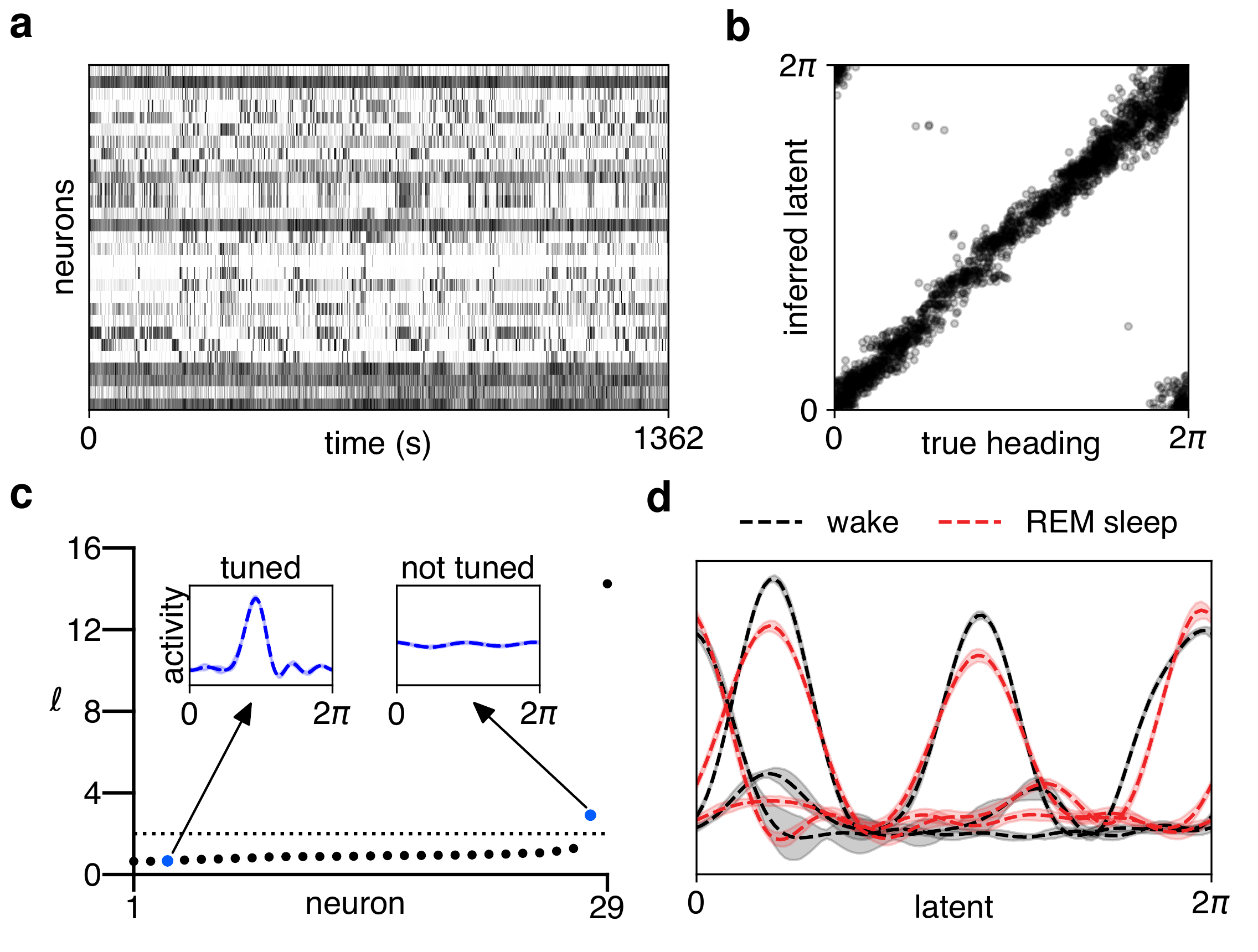}
        \end{flushleft}
    }
    \parbox{0.35\textwidth}{
    \caption{
    \label{fig:mouse}
    {\bfseries The mouse head direction circuit.}
    {\bfseries (a)}~
    Population activity recorded from mouse ADn during foraging.
    {\bfseries (b)}~
    Variational mean inferred by $T^1$-mGPLVM plotted against the true mouse head direction.
    {\bfseries (c)} Kernel length scales for the 29 neurons recorded.
    Dashed line: $\ell^2 = 4$ (maximum $d$ in the $T^1$-kernel).
    Insets: example neurons with low and high $\ell$.
    {\bfseries (d)} Tuning curves for three example neurons inferred during wake (black) and REM sleep (red).
    }
    }
    \vspace*{-2em}
\end{figure}

%% file: _sphere.tex
\begin{figure}[t!]
    \centering
    \includegraphics[width=0.8\textwidth]{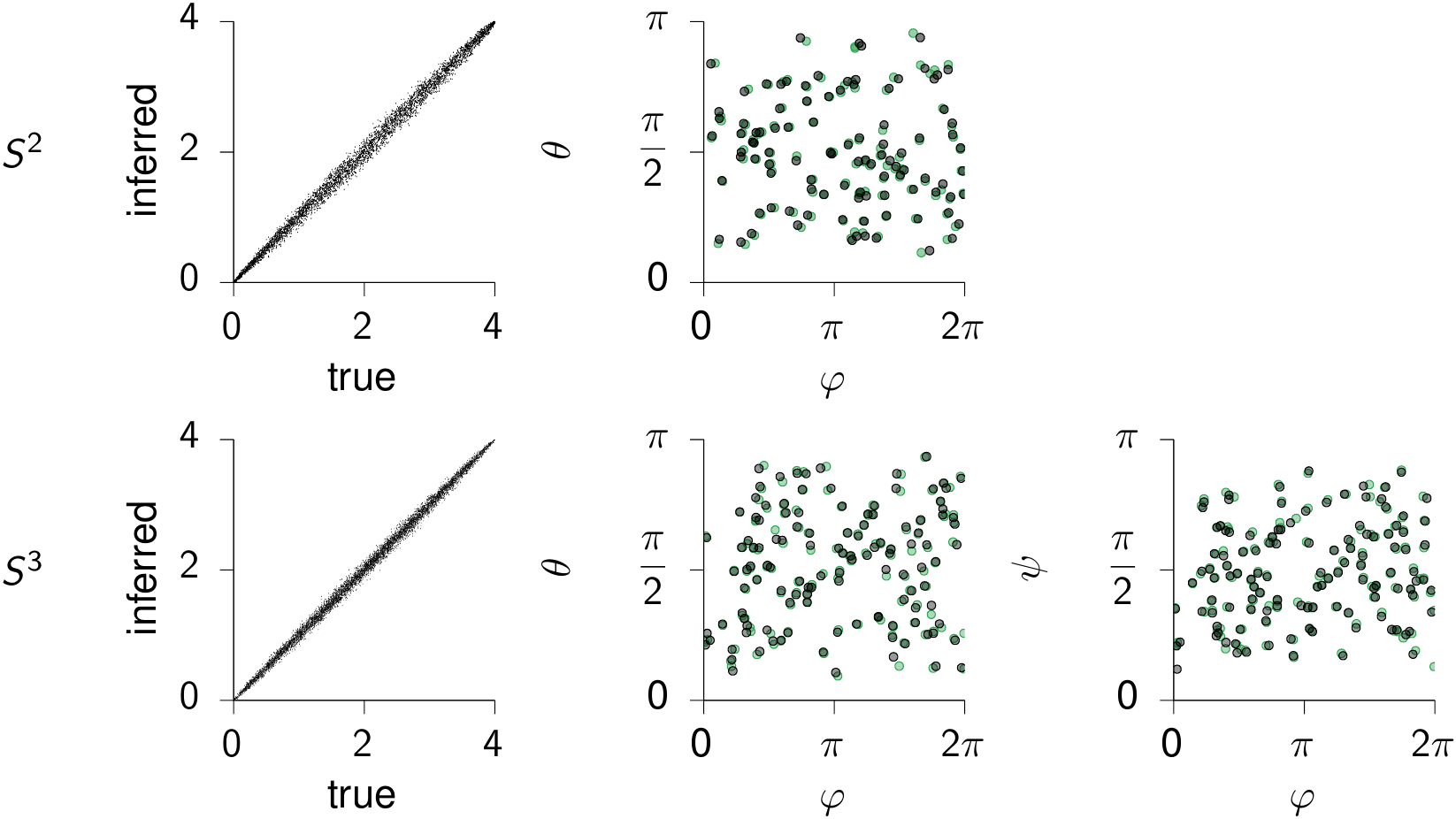}
    \caption{
        \label{fig:sphere}
        {\bfseries Applying mGPLVM to synthetic data on $S^2$ (top) and $S^3$ (bottom).}
        Pairwise distances between the variational means $\{ g^\mu_j \}$ are plotted against the corresponding pairwise distances between the true latent states $\{ g_j \}$ for $S^2$ (top left) and $S^3$ (bottom left).
        Since the log likelihood is a function of these pairwise distances through the kernel (\Cref{eq:k_Sn}), this illustrates that mGPLVM recovers the important features of the true latents.
        Inferred (black) and true (green) latent states in spherical coordinates for $S^2$ (top middle) and $S^3$ (bottom middle and bottom right).
        For $S^2$, we are showing the latent states in spherical polar coordinates $\b{g} = (\sin\theta\cos\varphi, \sin\theta\sin\varphi, \cos\theta)$ with $\theta \in [0, \pi]$ and $\varphi \in [0, 2\pi]$.
        For $S^3$, we use hyperspherical coordinates $\b{g} = (\sin\psi\sin\theta\cos\varphi, \sin\psi\sin\theta\sin\varphi, \sin\theta\cos\psi, \cos\theta)$ with $\theta, \psi \in [0, \pi]$ and $\varphi \in [0, 2\pi]$.
    }
\end{figure}

%% file: _ard.tex
\begin{figure}[t!]
    \centering
    \includegraphics[width=0.99\textwidth]{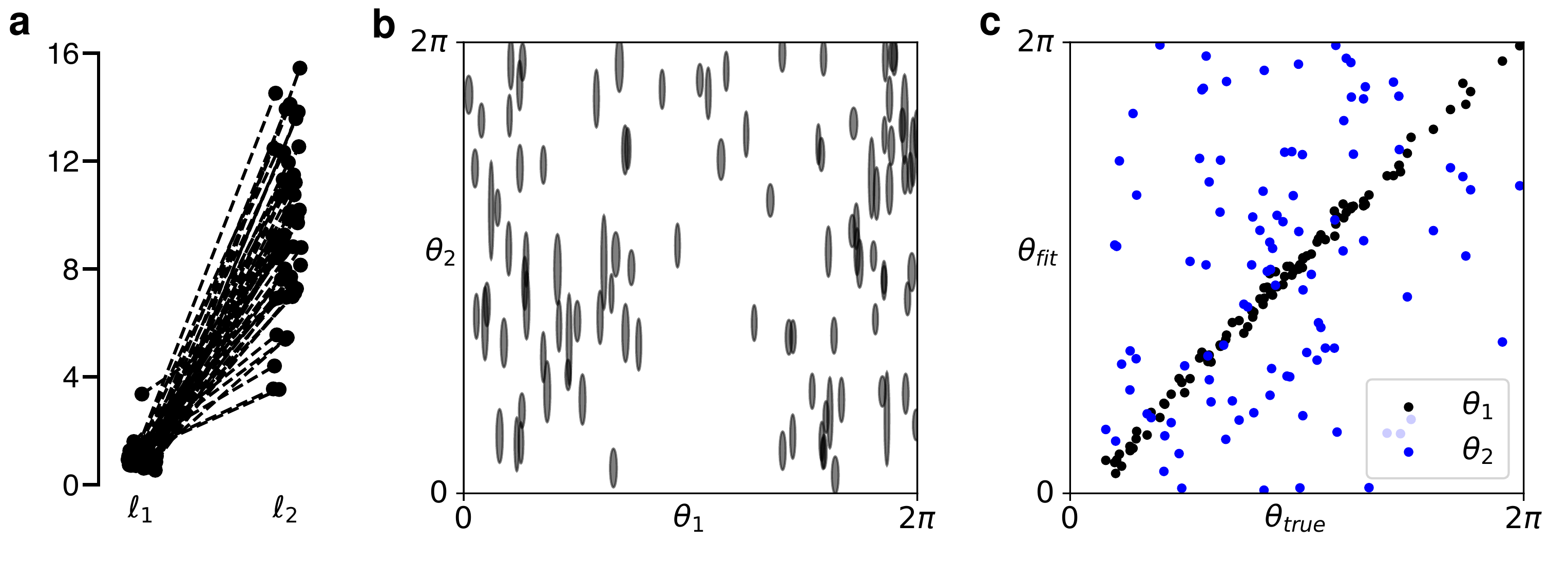}
    \caption{
        \label{fig:ard}
        {\bfseries Automatic relevance determination (ARD) in $T^2$-mGPLVM.}
	A $T^2$ model with ARD was fitted to the $T^1$ data in \Cref{fig:ring}.
	{\bfseries (a)} Length scales along each of the two dimensions for each neuron.
	{\bfseries (b)} Posterior variational distributions.
	Shading indicates $\pm 1$ s.t.d.\ around the posterior mean in each dimension.
	{\bfseries (c)} Variational mean plotted against the true latent state for each dimension.
    }
    \vspace*{-1em}
\end{figure}

%% file: _fly2d.tex
\begin{figure}[t!]
    \parbox{0.5\textwidth}{
    \includegraphics[width=0.47\textwidth]{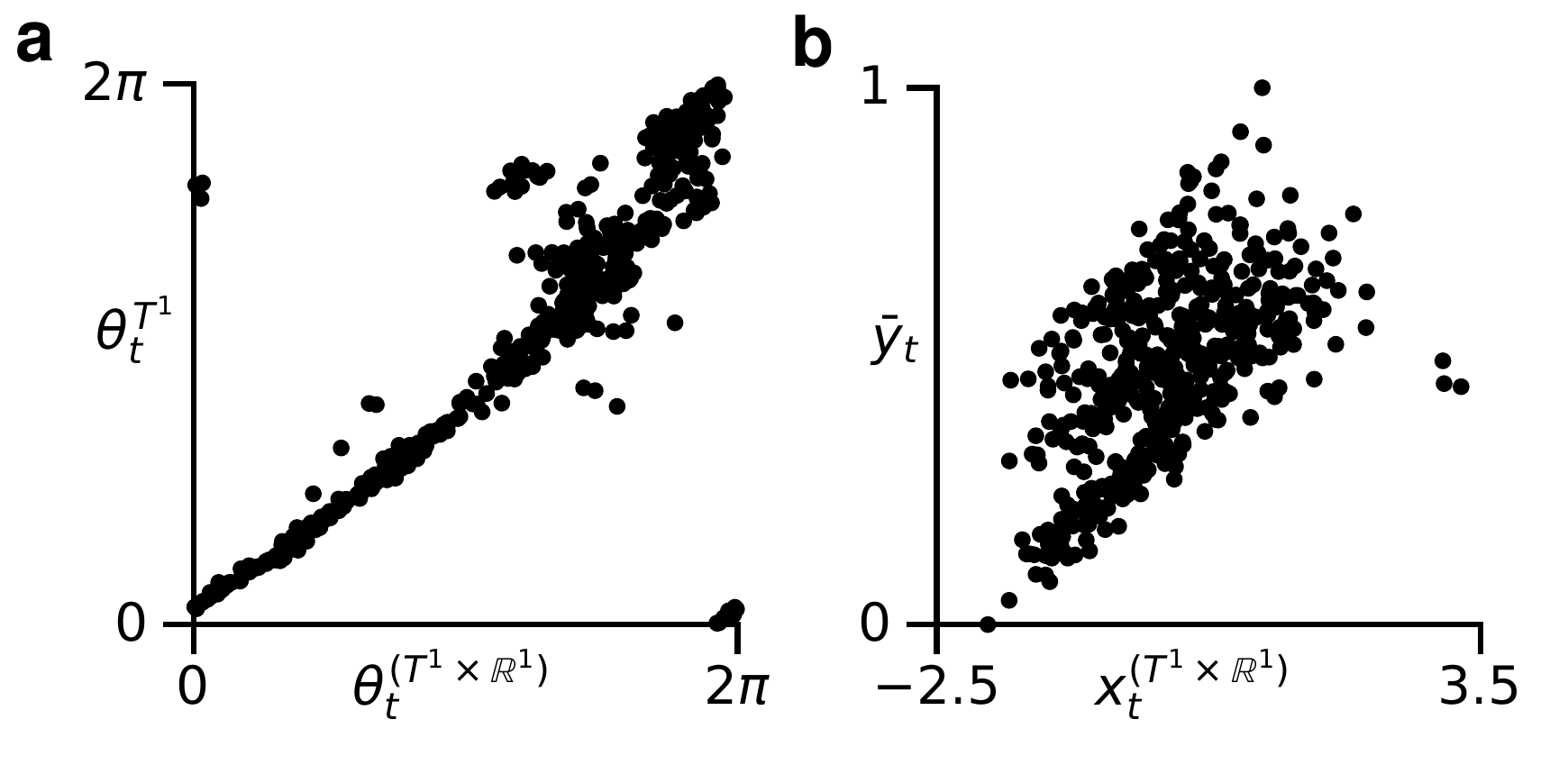}
    }
    \parbox{0.495\textwidth}{
    \caption{
        \label{fig:fly2d}
        {\bfseries $(T^1 \times \mathbb{R}^1)$-mGPLVM.}
        {\bfseries (a)}~Latent states inferred by $T^1$-mGPLVM (\Cref{fig:fly1d}a) against the periodic coordinate of a $(T^1 \times \mathbb{R}^1)$-mGPLVM fitted to the \textit{Drosophila} data.
        {\bfseries (b)}~Momentary average population activity $\bar{y}_t$ against the scalar Euclidean component of the $(T^1 \times \mathbb{R}^1)$ latent representation.
    }}
\end{figure}

%% file: manifold_GPLVM.bbl
\begin{thebibliography}{}

\bibitem[Chaudhuri et~al., 2019]{chaudhuri2019intrinsicAPP}
Chaudhuri, R., Gercek, B., Pandey, B., Peyrache, A., and Fiete, I. (2019).
\newblock The intrinsic attractor manifold and population dynamics of a
  canonical cognitive circuit across waking and sleep.
\newblock {\em Nature Neuroscience}, 22(9):1512--1520.

\bibitem[Davidson et~al., 2018]{Davidson2018APP}
Davidson, T.~R., Falorsi, L., De~Cao, N., Kipf, T., and Tomczak, J.~M. (2018).
\newblock Hyperspherical variational auto-encoders.
\newblock {\em 34th Conference on Uncertainty in Artificial Intelligence}.

\bibitem[Falorsi et~al., 2019]{Falorsi2019APP}
Falorsi, L., de~Haan, P., Davidson, T.~R., and Forr{\'{e}}, P. (2019).
\newblock {Reparameterizing Distributions on {Lie} Groups}.
\newblock {\em arXiv preprint arXiv:1903.02958}.

\bibitem[Jakob, 2012]{Jakob2012APP}
Jakob, W. (2012).
\newblock Numerically stable sampling of the von {Mises-Fisher} distribution on
  {$S^2$} (and other tricks).

\bibitem[Naesseth et~al., 2016]{Naesseth2016APP}
Naesseth, C.~A., Ruiz, F.~J., Linderman, S.~W., and Blei, D.~M. (2016).
\newblock Reparameterization gradients through acceptance-rejection sampling
  algorithms.
\newblock {\em arXiv preprint arXiv:1610.05683}.

\bibitem[Peyrache et~al., 2015a]{peyrache2015internallyAPP}
Peyrache, A., Lacroix, M.~M., Petersen, P.~C., and Buzs{\'a}ki, G. (2015a).
\newblock Internally organized mechanisms of the head direction sense.
\newblock {\em Nature Neuroscience}, 18(4):569--575.

\bibitem[Peyrache et~al., 2015b]{peyrache2015extracellularAPP}
Peyrache, A., Petersen, P., and Buzs{\'a}ki, G. (2015b).
\newblock Extracellular recordings from multi-site silicon probes in the
  anterior thalamus and subicular formation of freely moving mice.
\newblock {\em CRCNS.org}.
\newblock Dataset. https://doi.org/10.6080/K0G15XS1.

\bibitem[Rezende et~al., 2020]{Rezende2020APP}
Rezende, D.~J., Papamakarios, G., Racani{\`e}re, S., Albergo, M.~S., Kanwar,
  G., Shanahan, P.~E., and Cranmer, K. (2020).
\newblock Normalizing flows on tori and spheres.
\newblock {\em arXiv preprint arXiv:2022.02428}.

\bibitem[Rubin et~al., 2019]{rubin2019revealingAPP}
Rubin, A., Sheintuch, L., Brande-Eilat, N., Pinchasof, O., Rechavi, Y., Geva,
  N., and Ziv, Y. (2019).
\newblock Revealing neural correlates of behavior without behavioral
  measurements.
\newblock {\em Nature communications}, 10:1--14.

\bibitem[Sola et~al., 2018]{Sola2018APP}
Sola, J., Deray, J., and Atchuthan, D. (2018).
\newblock A micro {Lie} theory for state estimation in robotics.
\newblock {\em arXiv preprint arXiv:1812.01537}.

\bibitem[Straub, 2017]{Straub2017APP}
Straub, J. (2017).
\newblock Bayesian inference with the {von-Mises-Fisher} distribution in 3d.

\bibitem[Titsias, 2009]{Titsias2009APP}
Titsias, M.~K. (2009).
\newblock {Variational learning of inducing variables in sparse Gaussian
  processes}.
\newblock In {\em Journal of Machine Learning Research}, volume~5, pages
  567--574.

\bibitem[Ulrich, 1984]{Ulrich1984APP}
Ulrich, G. (1984).
\newblock Computer generation of distributions on the {M-sphere}.
\newblock {\em Journal of the Royal Statistical Society: Series C (Applied
  Statistics)}, 33(2):158--163.

\end{thebibliography}


\begin{thebibliography}{}

\bibitem[Arvanitidis et~al., 2017]{arvanitidis2017latent}
Arvanitidis, G., Hansen, L.~K., and Hauberg, S. (2017).
\newblock Latent space oddity: on the curvature of deep generative models.
\newblock {\em arXiv preprint arXiv:1710.11379}.

\bibitem[Banino et~al., 2018]{Banino2018}
Banino, A., Barry, C., Uria, B., Blundell, C., Lillicrap, T., Mirowski, P.,
  Pritzel, A., Chadwick, M.~J., Degris, T., Modayil, J., Wayne, G., Soyer, H.,
  Viola, F., Zhang, B., Goroshin, R., Rabinowitz, N., Pascanu, R., Beattie, C.,
  Petersen, S., Sadik, A., Gaffney, S., King, H., Kavukcuoglu, K., Hassabis,
  D., Hadsell, R., and Kumaran, D. (2018).
\newblock {Vector-based navigation using grid-like representations in
  artificial agents}.
\newblock {\em Nature}, 557(7705):429--433.

\bibitem[Borovitskiy et~al., 2020]{borovitskiy2020matern}
Borovitskiy, V., Terenin, A., Mostowsky, P., and Deisenroth, M.~P. (2020).
\newblock Mat{\'e}rn {Gaussian} processes on {Riemannian} manifolds.
\newblock {\em arXiv preprint arXiv:2006.10160}.

\bibitem[Burda et~al., 2015]{burda2015importance}
Burda, Y., Grosse, R., and Salakhutdinov, R. (2015).
\newblock Importance weighted autoencoders.
\newblock {\em arXiv preprint arXiv:1509.00519}.

\bibitem[Chaudhuri et~al., 2019]{Chaudhuri2019}
Chaudhuri, R., Ger{\c{c}}ek, B., Pandey, B., Peyrache, A., and Fiete, I.
  (2019).
\newblock {The intrinsic attractor manifold and population dynamics of a
  canonical cognitive circuit across waking and sleep}.
\newblock {\em Nature Neuroscience}, 22(9):1512--1520.

\bibitem[Churchland and Shenoy, 2007]{churchland2007temporal}
Churchland, M.~M. and Shenoy, K.~V. (2007).
\newblock Temporal complexity and heterogeneity of single-neuron activity in
  premotor and motor cortex.
\newblock {\em Journal of neurophysiology}, 97:4235--4257.

\bibitem[Constantinescu et~al., 2016]{constantinescu2016organizing}
Constantinescu, A.~O., O’Reilly, J.~X., and Behrens, T.~E. (2016).
\newblock Organizing conceptual knowledge in humans with a gridlike code.
\newblock {\em Science}, 352:1464--1468.

\bibitem[Cunningham and Byron, 2014]{Cunningham2014}
Cunningham, J.~P. and Byron, M.~Y. (2014).
\newblock Dimensionality reduction for large-scale neural recordings.
\newblock {\em Nature Neuroscience}, 17(11):1500--1509.

\bibitem[Cunningham and Ghahramani, 2015]{cunningham2015linear}
Cunningham, J.~P. and Ghahramani, Z. (2015).
\newblock Linear dimensionality reduction: Survey, insights, and
  generalizations.
\newblock {\em The Journal of Machine Learning Research}, 16:2859--2900.

\bibitem[Davidson et~al., 2018]{Davidson2018}
Davidson, T.~R., Falorsi, L., De~Cao, N., Kipf, T., and Tomczak, J.~M. (2018).
\newblock Hyperspherical variational auto-encoders.
\newblock {\em 34th Conference on Uncertainty in Artificial Intelligence}.

\bibitem[Elgammal and Lee, 2008]{elgammal2008tracking}
Elgammal, A. and Lee, C.-S. (2008).
\newblock Tracking people on a torus.
\newblock {\em IEEE Transactions on Pattern Analysis and Machine Intelligence},
  31(3):520--538.

\bibitem[Falorsi et~al., 2019]{Falorsi2019}
Falorsi, L., de~Haan, P., Davidson, T.~R., and Forr{\'{e}}, P. (2019).
\newblock Reparameterizing distributions on {Lie} groups.
\newblock {\em arXiv preprint arXiv:1903.02958}.

\bibitem[Falorsi and Forr{\'e}, 2020]{falorsi2020neural}
Falorsi, L. and Forr{\'e}, P. (2020).
\newblock Neural ordinary differential equations on manifolds.
\newblock {\em arXiv preprint arXiv:2006.06663}.

\bibitem[Feragen et~al., 2015]{feragen2015geodesic}
Feragen, A., Lauze, F., and Hauberg, S. (2015).
\newblock Geodesic exponential kernels: When curvature and linearity conflict.
\newblock In {\em Proceedings of the IEEE Conference on Computer Vision and
  Pattern Recognition}, pages 3032--3042.

\bibitem[Finkelstein et~al., 2015]{Finkelstein2015}
Finkelstein, A., Derdikman, D., Rubin, A., Foerster, J.~N., Las, L., and
  Ulanovsky, N. (2015).
\newblock {Three-dimensional head-direction coding in the bat brain}.
\newblock {\em Nature}, 517(7533):159--164.

\bibitem[Hafting et~al., 2005]{hafting2005microstructure}
Hafting, T., Fyhn, M., Molden, S., Moser, M.-B., and Moser, E.~I. (2005).
\newblock Microstructure of a spatial map in the entorhinal cortex.
\newblock {\em Nature}, 436:801--806.

\bibitem[Hardcastle et~al., 2017]{hardcastle2017multiplexed}
Hardcastle, K., Maheswaranathan, N., Ganguli, S., and Giocomo, L.~M. (2017).
\newblock A multiplexed, heterogeneous, and adaptive code for navigation in
  medial entorhinal cortex.
\newblock {\em Neuron}, 94:375--387.

\bibitem[Jacob et~al., 2017]{jacob2017independent}
Jacob, P.-Y., Casali, G., Spieser, L., Page, H., Overington, D., and Jeffery,
  K. (2017).
\newblock An independent, landmark-dominated head-direction signal in
  dysgranular retrosplenial cortex.
\newblock {\em Nature neuroscience}, 20(2):173--175.

\bibitem[Jayasumana et~al., 2015]{Jayasumana2015}
Jayasumana, S., Hartley, R., Salzmann, M., Li, H., and Harandi, M. (2015).
\newblock Kernel methods on {Riemannian} manifolds with {Gaussian} {RBF}
  kernels.
\newblock {\em IEEE transactions on pattern analysis and machine intelligence},
  37(12):2464--2477.

\bibitem[Kingma and Ba, 2014]{kingma2014adam}
Kingma, D.~P. and Ba, J. (2014).
\newblock Adam: A method for stochastic optimization.
\newblock {\em arXiv preprint arXiv:1412.6980}.

\bibitem[Kingma and Welling, 2014]{Kingma2014}
Kingma, D.~P. and Welling, M. (2014).
\newblock Auto-encoding variational {Bayes}.
\newblock In {\em 2nd International Conference on Learning Representations,
  ICLR 2014}.

\bibitem[Lawrence, 2005]{Lawrence2005}
Lawrence, N. (2005).
\newblock {Probabilistic non-linear principal component analysis with Gaussian
  process latent variable models}.
\newblock {\em Journal of Machine Learning Research}, 6:1783--1816.

\bibitem[Lou et~al., 2020]{lou2020neural}
Lou, A., Lim, D., Katsman, I., Huang, L., Jiang, Q., Lim, S.-N., and De~Sa, C.
  (2020).
\newblock Neural manifold ordinary differential equations.
\newblock {\em arXiv preprint arXiv:2006.10254}.

\bibitem[Maaten and Hinton, 2008]{maaten2008visualizing}
Maaten, L. v.~d. and Hinton, G. (2008).
\newblock Visualizing data using t-{SNE}.
\newblock {\em Journal of machine learning research}, 9:2579--2605.

\bibitem[MacKay, 1998]{MacKay1998}
MacKay, D.~J. (1998).
\newblock Introduction to {Gaussian} processes.
\newblock {\em NATO ASI series. Series F: computer and system sciences}, pages
  133--165.

\bibitem[Mallasto and Feragen, 2018]{Mallasto2018wrapped}
Mallasto, A. and Feragen, A. (2018).
\newblock Wrapped {Gaussian} process regression on {Riemannian} manifolds.
\newblock In {\em Proceedings of the IEEE Conference on Computer Vision and
  Pattern Recognition}, pages 5580--5588.

\bibitem[Mallasto et~al., 2019]{mallasto2019probabilistic}
Mallasto, A., Hauberg, S., and Feragen, A. (2019).
\newblock Probabilistic {Riemannian} submanifold learning with wrapped
  {Gaussian} process latent variable models.
\newblock In {\em The 22nd International Conference on Artificial Intelligence
  and Statistics}, pages 2368--2377.

\bibitem[Masullo et~al., 2019]{masullo2019genetically}
Masullo, L., Mariotti, L., Alexandre, N., Freire-Pritchett, P., Boulanger, J.,
  and Tripodi, M. (2019).
\newblock Genetically defined functional modules for spatial orienting in the
  mouse superior colliculus.
\newblock {\em Current Biology}, 29:2892--2904.

\bibitem[Mathieu and Nickel, 2020]{mathieu2020riemannian}
Mathieu, E. and Nickel, M. (2020).
\newblock Riemannian continuous normalizing flows.
\newblock {\em arXiv preprint arXiv:2006.10605}.

\bibitem[Navarro et~al., 2017]{navarro2017multivariate}
Navarro, A.~K., Frellsen, J., and Turner, R.~E. (2017).
\newblock The multivariate generalised von mises distribution: inference and
  applications.
\newblock In {\em Thirty-First AAAI Conference on Artificial Intelligence}.

\bibitem[Neal, 2012]{Neal2012}
Neal, R.~M. (2012).
\newblock {\em Bayesian learning for neural networks}, volume 118.
\newblock Springer.

\bibitem[Pandarinath et~al., 2018]{Pandarinath2018}
Pandarinath, C., O'Shea, D.~J., Collins, J., Jozefowicz, R., Stavisky, S.~D.,
  Kao, J.~C., Trautmann, E.~M., Kaufman, M.~T., Ryu, S.~I., Hochberg, L.~R.,
  Henderson, J.~M., Shenoy, K.~V., Abbott, L.~F., and Sussillo, D. (2018).
\newblock {Inferring single-trial neural population dynamics using sequential
  auto-encoders}.
\newblock {\em Nature Methods}, 15(10):805--815.

\bibitem[Peyrache et~al., 2015]{peyrache2015extracellular}
Peyrache, A., Petersen, P., and Buzs{\'a}ki, G. (2015).
\newblock Extracellular recordings from multi-site silicon probes in the
  anterior thalamus and subicular formation of freely moving mice.
\newblock {\em CRCNS.org}.
\newblock Dataset. https://doi.org/10.6080/K0G15XS1.

\bibitem[Rasmussen and Williams, 2006]{williams2006gaussian}
Rasmussen, C.~E. and Williams, C.~K. (2006).
\newblock {\em Gaussian processes for machine learning}.
\newblock MIT press Cambridge, MA.

\bibitem[Rey et~al., 2019]{rey2019diffusion}
Rey, L. A.~P., Menkovski, V., and Portegies, J.~W. (2019).
\newblock Diffusion variational autoencoders.
\newblock {\em arXiv preprint arXiv:1901.08991}.

\bibitem[Rezende et~al., 2014]{Rezende2014}
Rezende, D.~J., Mohamed, S., and Wierstra, D. (2014).
\newblock {Stochastic backpropagation and approximate inference in deep
  generative models}.
\newblock In {\em 31st International Conference on Machine Learning, ICML
  2014}, pages 3057--3070.

\bibitem[Rezende et~al., 2020]{Rezende2020}
Rezende, D.~J., Papamakarios, G., Racani{\`e}re, S., Albergo, M.~S., Kanwar,
  G., Shanahan, P.~E., and Cranmer, K. (2020).
\newblock Normalizing flows on tori and spheres.
\newblock {\em arXiv preprint arXiv:2022.02428}.

\bibitem[Rubin et~al., 2019]{rubin2019revealing}
Rubin, A., Sheintuch, L., Brande-Eilat, N., Pinchasof, O., Rechavi, Y., Geva,
  N., and Ziv, Y. (2019).
\newblock Revealing neural correlates of behavior without behavioral
  measurements.
\newblock {\em Nature communications}, 10:1--14.

\bibitem[Seelig and Jayaraman, 2015]{Seelig2015}
Seelig, J.~D. and Jayaraman, V. (2015).
\newblock {Neural dynamics for landmark orientation and angular path
  integration}.
\newblock {\em Nature}, 521(7551):186--191.

\bibitem[Shepard and Metzler, 1971]{shepard1971mental}
Shepard, R.~N. and Metzler, J. (1971).
\newblock Mental rotation of three-dimensional objects.
\newblock {\em Science}, 171:701--703.

\bibitem[Sola et~al., 2018]{Sola2018}
Sola, J., Deray, J., and Atchuthan, D. (2018).
\newblock A micro {Lie} theory for state estimation in robotics.
\newblock {\em arXiv preprint arXiv:1812.01537}.

\bibitem[Stringer et~al., 2019]{Stringer2019}
Stringer, C., Pachitariu, M., Steinmetz, N., Carandini, M., and Harris, K.~D.
  (2019).
\newblock {High-dimensional geometry of population responses in visual cortex}.
\newblock {\em Nature}, 571(7765):361--365.

\bibitem[Sussillo and Barak, 2013]{sussillo2013opening}
Sussillo, D. and Barak, O. (2013).
\newblock Opening the black box: low-dimensional dynamics in high-dimensional
  recurrent neural networks.
\newblock {\em Neural computation}, 25:626--649.

\bibitem[Titsias, 2009]{Titsias2009}
Titsias, M.~K. (2009).
\newblock {Variational learning of inducing variables in sparse Gaussian
  processes}.
\newblock In {\em Journal of Machine Learning Research}, volume~5, pages
  567--574.

\bibitem[Titsias and Lawrence, 2010]{Titsias2010a}
Titsias, M.~K. and Lawrence, N.~D. (2010).
\newblock {Bayesian Gaussian process latent variable model}.
\newblock In {\em Journal of Machine Learning Research}, volume~9, pages
  844--851.

\bibitem[Tosi et~al., 2014]{tosi2014metrics}
Tosi, A., Hauberg, S., Vellido, A., and Lawrence, N.~D. (2014).
\newblock Metrics for probabilistic geometries.
\newblock {\em arXiv preprint arXiv:1411.7432}.

\bibitem[Turner-Evans, 2020]{turner2020kir}
Turner-Evans, D.~B. (2020).
\newblock Kir.zip.
\newblock {\em Janelia Research Campus}.
\newblock Dataset. https://doi.org/10.25378/janelia.12490325.v1.

\bibitem[Turner-Evans et~al., 2020]{turner2020neuroanatomical}
Turner-Evans, D.~B., Jensen, K.~T., Ali, S., Paterson, T., Sheridan, A., Ray,
  R.~P., Wolff, T., Lauritzen, J.~S., Rubin, G.~M., Bock, D.~D., and Jayaraman,
  V. (2020).
\newblock The neuroanatomical ultrastructure and function of a biological ring
  attractor.
\newblock {\em Neuron}, 108:145--163.

\bibitem[Urtasun et~al., 2008]{urtasun2008topologically}
Urtasun, R., Fleet, D.~J., Geiger, A., Popovi{\'c}, J., Darrell, T.~J., and
  Lawrence, N.~D. (2008).
\newblock Topologically-constrained latent variable models.
\newblock In {\em Proceedings of the 25th international conference on Machine
  learning}, pages 1080--1087.

\bibitem[Wang and Wang, 2019]{wang2019riemannian}
Wang, P.~Z. and Wang, W.~Y. (2019).
\newblock Riemannian normalizing flow on variational {Wasserstein} autoencoder
  for text modeling.
\newblock {\em arXiv preprint arXiv:1904.02399}.

\bibitem[Wilson et~al., 2018]{Wilson2018}
Wilson, J.~J., Alexandre, N., Trentin, C., and Tripodi, M. (2018).
\newblock Three-dimensional representation of motor space in the mouse superior
  colliculus.
\newblock {\em Current Biology}, 28(11):1744--1755.e12.

\bibitem[Wu et~al., 2018]{Wu2018}
Wu, A., Pashkovski, S., Datta, S.~R., and Pillow, J.~W. (2018).
\newblock Learning a latent manifold of odor representations from neural
  responses in piriform cortex.
\newblock In {\em Advances in Neural Information Processing Systems}, pages
  5378--5388.

\bibitem[Wu et~al., 2017]{Wu2017}
Wu, A., Roy, N.~A., Keeley, S., and Pillow, J.~W. (2017).
\newblock Gaussian process based nonlinear latent structure discovery in
  multivariate spike train data.
\newblock In {\em Advances in Neural Information Processing Systems}, pages
  3497--3506.

\bibitem[Yu et~al., 2009]{Yu2009}
Yu, B.~M., Cunningham, J.~P., Santhanam, G., Ryu, S.~I., Shenoy, K.~V., and
  Sahani, M. (2009).
\newblock Gaussian-process factor analysis for low-dimensional single-trial
  analysis of neural population activity.
\newblock {\em Journal of Neurophysiology}, 102(1):614--635.

\end{thebibliography}
